\documentclass{bmvc2k}

\title{Circumventing Magnitude Collapse via\\ Modular Two-Stage Open-Set Recognition}

\addauthor{Laurin van den Bergh}{bergh@ifi.uzh.ch}{1,2}
\addauthor{Halil Bisgin}{bisgin@umich.edu}{3}
\addauthor{Andres Palechor}{andrespalechor11@gmail.com}{1,\dag}
\addauthor{Mike Suter}{mi.suter@protonmail.com}{1,\dag}
\addauthor{Manuel G\"unther}{siebenkopf@googlemail.com}{1,2,\dag}

\addinstitution{
    Department of Informatics\\
    University of Zurich\\
    Zurich, Switzerland
}
\addinstitution{
    Digital Society Initiative\\
    University of Zurich\\
    Zurich, Switzerland
}
\addinstitution{
    College of Innovation \& Technology\\
    University of Michigan-Flint\\
    Michigan, USA
}

\runninghead{van den Bergh et al.}{Modular Two-Stage Open-Set Recognition}

\def\eg{\emph{e.g}\bmvaOneDot}

\newif\ifpreprint
\preprinttrue

\usepackage{url}
\usepackage{hyperref}
\usepackage{sidecap}
\sidecaptionvpos{figure}{t}
\usepackage{lmodern}
\usepackage{graphicx}
\usepackage{multirow,tabularx}
\usepackage{float}
\usepackage{todonotes}
\usepackage{textcomp}
\usepackage{amsmath,amssymb}
\usepackage{ifthen}
\usepackage{color}
\usepackage{xcolor}
\usepackage{tabularx}
\usepackage{booktabs}
\usepackage{longtable}
\usepackage{footmisc}
\usepackage{rotating}
\usepackage{enumitem}

\makeatletter
\let\c@subfigure\@undefined
\let\subfigure\@undefined
\let\c@subtable\@undefined
\let\subtable\@undefined
\makeatother
\usepackage[labelformat=parens,subrefformat=parens]{subcaption}  
\usepackage{hyperref}

\DeclareGraphicsExtensions{.pdf,.jpg,.png}
\graphicspath{{./graphics/}{./figures/}}

\newcolumntype{C}{X<{\centering}}

\definecolor{lightgreen}{rgb}{0.67, 0.88, 0.69}
\definecolor{darkgreen}{rgb}{0,0.55,0}
\definecolor{linkcolor}{rgb}{0,0,.65}

\newcommand\Caption[3][]{%
  \caption[#2]{%
    \ifstrempty{#1}{}{\label{#1}}%
    \textsc{#2}. \it #3%
  }%
}

\newcommand\ie{\textit{i.\,e.}}

\renewcommand\sec[1]{Section~\ref{sec:#1}}
\newcommand\fig[1]{Figure~\ref{fig:#1}}
\newcommand\sfig[1]{\fig{#1}}
\newcommand\tab[1]{Table~\ref{tab:#1}}

\newcommand\apx[1]{Appendix~\ref{sec:#1}}

\newcommand\T{\ensuremath{^\mathrm{T}}}

\DeclareMathOperator*{\argmax}{arg\,max}

\newcommand{\p}[1]{\ensuremath{P_{#1}}}

\newcommand{\knowns}{\ensuremath{\mathcal{K}}}
\newcommand{\knownlabels}{\ensuremath{\mathcal{Y}}}
\newcommand{\aux}{\ensuremath{\mathcal{A}}}
\newcommand{\unknowns}{\ensuremath{\mathcal{U}}}

\newcommand{\threshold}{\ensuremath{\theta}}
\newcommand{\fprval}{\ensuremath{\zeta}}

\newcommand{\metric}{\ensuremath{\mathcal{M}}}
\newcommand{\rl}{\ensuremath{f}} 
\newcommand{\pp}{\ensuremath{g}} 
\newcommand{\comb}[2]{\ensuremath{\pp{}_\textrm{#2} \circ \rl{}_\textrm{#1}}} 
\newcommand{\deltarl}[2][]{\ensuremath{\Delta_{\rl{}}(\rl{}_{\mathrm{#2}} \mid \pp{}_{\mathrm{#1}})}}
\newcommand{\deltapp}[2][]{\ensuremath{\Delta_{\pp{}}(\pp{}_{\mathrm{#2}} \mid \rl{}_{\mathrm{#1}})}}

\newcommand{\sample}{\ensuremath{\mathbf x}} 
\newcommand{\backbone}{\ensuremath{F}} 
\newcommand{\df}{\ensuremath{\mathbf \varphi}} 
\newcommand{\mav}{\ensuremath{\mu}} 
\newcommand{\logitW}{\ensuremath{\mathbf W}} 
\newcommand{\logits}{\ensuremath{\mathbf z}}
\newcommand{\confidence}{\ensuremath{\mathbf y}} 

\newcommand{\loss}{\ensuremath{\mathcal{J}}} 
\newcommand{\clabel}{\ensuremath{\tau}} 
\newcommand{\pred}{\ensuremath{k^{*}}}
\newcommand{\score}{\ensuremath{\gamma}^{*}}
\newcommand{\indicator}{\ensuremath{\mathbb{E}}} 
\newcommand{\representation}{\ensuremath{\mathcal{R}}} 
\newcommand{\scores}{\ensuremath{\mathcal{P}}} 

\newcommand{\lb}{\ensuremath{l}} 

\newcommand{\decision}{G}
\newcommand{\dfangle}{\ensuremath{\alpha}} 

\newcommand\garbage{$K+1$ class}

\makeatletter
\renewcommand{\citet}[1]{%
  \@ifundefined{b@#1\@extra@b@citeb}%
    {\textcolor{black}{??}}%
    {\begingroup\NAT@parse{#1}\textcolor{black}{\NAT@name}\endgroup}%
  ~\cite{#1}%
}
\makeatother
\renewcommand{\citep}[1]{\cite{#1}}

\begin{document}

\maketitle
\makeatletter
\BMVA@blfootnote{$^\dagger$Work done while at the University of Zurich.}
\makeatother
\ifpreprint
\AddToShipoutPicture*{%
  \put(0,\LenToUnit{\paperheight-1.1cm}){%
    \makebox(\LenToUnit{\paperwidth},0){\small\sffamily\textbf{Preprint.} Accepted for publication at the British Machine Vision Conference (BMVC) 2026.}}%
}
\fi

\begin{abstract}



Open-set recognition (OSR) requires neural networks to classify known classes while rejecting unknown samples, which is critical for real-world deployment.
So far, OSR research studied and developed representation learning and postprocessing methods independently and their interactions remain unexplored.
In this paper, we reconsider OSR as a modular two-stage pipeline and present the first systematic study of interactions between representation learning and postprocessing methods across dataset scales and usage of known unknown data.
First, we discover a failure mode we term \emph{magnitude collapse}, where representation learning methods that utilize known unknown data can suffer performance degradation at large scale and irreversibly destroy discriminative information, despite excelling at small scale.
Second, we study the modular interactions and reveal when they can be leveraged for modular performance gains via two-stage processing and where interaction effects amplify performance degradation due to magnitude collapse.
Third, we show how these findings can be used to achieve state-of-the-art performance with a simple $K+1$ class classifier  baseline and magnitude-aware postprocessing while circumventing magnitude collapse.
Finally, our results demonstrate that small-scale evaluations with known unknown data are not predictive of large-scale performance, invalidating current best practices in OSR research.
Code available at \url{https://github.com/TwoDigitsOneNumber/Two-Stage-OSR}.

\end{abstract}

\section{Introduction}
\label{sec:introduction}


The rapid advancement of deep learning methods for image recognition increasingly promotes their real-world adoption, which requires them to adequately detect and handle unknown inputs for the reliability and safety of these systems \citep{scheirer2013toward,hendrycks2017baseline,vaze2022openset}.
This task is typically studied under the two closely-related problem formulations: \emph{Open-set Recognition} (OSR) and \emph{Out-of-Distribution} (OOD) detection.
Both aim to improve the robustness of classifiers by detecting test-time distributional shifts.

While the categorization of OSR methods into \emph{representation learning} and \emph{postprocessing} methods is commonly understood, current OSR methods are studied in isolation or compared as standalone methods, neglecting their differences and modular nature and potential for improvements through combinations.
Since the \emph{interaction effects} between representation learning and postprocessors have neither been explored nor formalized for OSR methods, we ask: can representation learning methods enhance or shape feature representations to amplify downstream postprocessors performance?
Or vice versa: is the optimal choice of postprocessor dependent on the representation learning training objective?









\newcommand{\legoblock}[7]{
  \pgfmathsetmacro{\studheight}{#7}
  \pgfmathsetmacro{\studwidth}{#4/5}
  \pgfmathsetmacro{\studspacing}{(#4 - 2*\studwidth)/3}
  \pgfmathsetmacro{\firstStudY}{#2 + \studspacing}
  \pgfmathsetmacro{\secondStudY}{#2 + \studspacing + \studwidth + \studspacing}
  \pgfmathsetmacro{\centerX}{#1 + (#3-\studheight)/2}
  \pgfmathsetmacro{\centerY}{#2 + #4/2}
  \pgfmathsetmacro{\studX}{#1 + #3-\studheight}
  \pgfmathsetmacro{\highlightY}{#2 + #4/2}

  \draw[thick, fill=#5] (#1,#2) rectangle (#1+#3-\studheight,#2+#4);

  \draw[thick, fill=#5] (\studX,\firstStudY) rectangle (\studX+\studheight,\firstStudY+\studwidth);

  \draw[thick, fill=#5] (\studX,\secondStudY) rectangle (\studX+\studheight,\secondStudY+\studwidth);

  \node[align=center, text=black, rotate=270] at (\centerX,\centerY) {#6};
}

\begin{figure*}[t]

    \centering
    \resizebox{1.02\textwidth}{!}{
    \begin{tikzpicture}[
        input/.style={
            rectangle,
            draw=gray,
            thick,
            minimum width=0.8cm,
            minimum height=0.8cm,
            fill=gray!20,
            font=\tiny\bf
        },
        arrow/.style={
            ->,
            thick,
            >=stealth
        },
        grouplabel/.style={
            font=\large\bf,
            text centered
        }
    ]

    \def\brickheight{0.8}
    \def\brickwidth{1.2}
    \def\studheight{0.15}

    \def\inputxknown{-3}
    \def\inputxaux{-2}
    \def\inputxunknown{-1}
    \def\reprx{.5}
    \def\postx{2.7}
    \def\scoresx{4.4}

    \def\textscaling{.5}
    \def\legotextscaling{.7}

    \def\arrowsy{1.6}
    \def\bottombricky{-0.8}

    \def\margin{0.15}

    \definecolor{reprcolor}{HTML}{81ff81}  
    \definecolor{mmcolor}{HTML}{ff6568}  
    \definecolor{postcolor}{HTML}{FFFF47}  
    \definecolor{macolor}{HTML}{79c9ff}  

    \def\reprcolor{reprcolor}
    \def\postcolor{postcolor}
    \def\mmcolor{mmcolor}
    \def\macolor{macolor}



    \node[grouplabel, align=center, rotate=90] at (\inputxknown,   3.7) {\scalebox{\textscaling}{Known}};
    \node[grouplabel, align=center, rotate=90] at (\inputxaux,     3.7) {\scalebox{\textscaling}{Known Unknown}};
    \node[grouplabel, align=center, rotate=90] at (\inputxunknown, 3.7) {\scalebox{\textscaling}{Unknown}};

    \node[grouplabel, align=center] (input_label) at (\inputxaux, 1.35) {\scalebox{\textscaling}{CIFAR+10}\\[-5pt]\scalebox{\textscaling}{(low similarity)}};
    \node (small1) at (\inputxknown,   2.2) {\includegraphics[width=1.0cm]{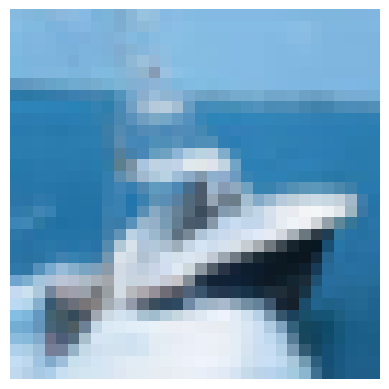}};
    \node (small2) at (\inputxaux,     2.2) {\includegraphics[width=1.0cm]{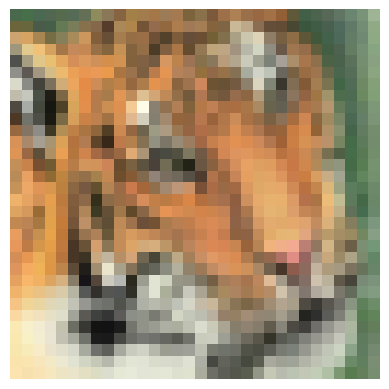}};
    \node (small3) at (\inputxunknown, 2.2) {\includegraphics[width=1.0cm]{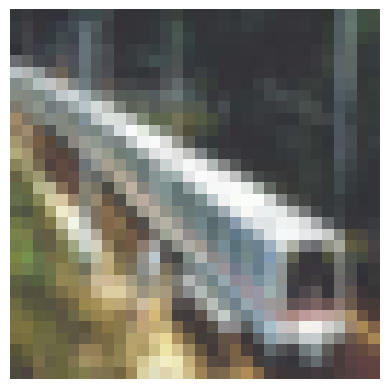}};

    \node[grouplabel, align=center] (input_label) at (\inputxaux, -0.4) {\scalebox{\textscaling}{ImageNet $P_3$}\\[-5pt]\scalebox{\textscaling}{(high similarity)}};
    \node (large1) at (\inputxknown,   0.5) {\includegraphics[width=1.0cm]{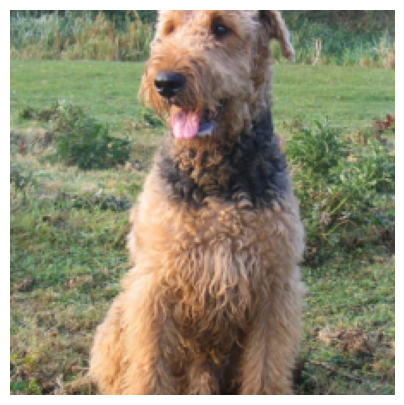}};
    \node (large2) at (\inputxaux,     0.5) {\includegraphics[width=1.0cm]{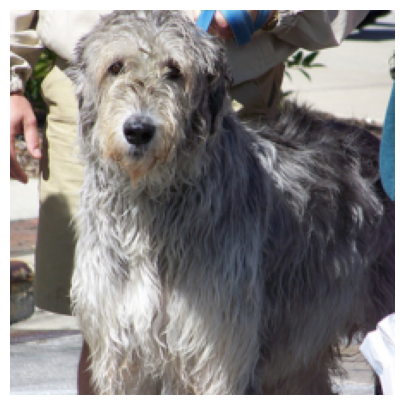}};
    \node (large3) at (\inputxunknown, 0.5) {\includegraphics[width=1.0cm]{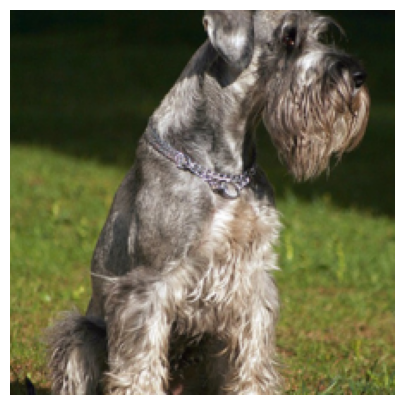}};

    \draw[arrow] (\inputxunknown+0.8, \arrowsy) -- (\reprx-\margin, \arrowsy);


    \node[grouplabel, align=center] (repr_label) at (\reprx+\brickheight/2, 4.0) {\scalebox{\textscaling}{Representation}\\[-5pt]\scalebox{\textscaling}{Learning}};
    \foreach \y/\label in {{\arrowsy+\brickwidth/2+\margin}/CE, {\arrowsy-\brickwidth/2}/\garbage{}} {
        \legoblock{\reprx}{\y}{\brickheight}{\brickwidth}{\reprcolor}{\scalebox{\legotextscaling}{\label}}{\studheight}
    }
    \node at ({\reprx+(\brickheight-\studheight)/2}, 0.8) {\scalebox{1.1}{$\mathbf{\vdots}$}};
    \legoblock{\reprx}{\bottombricky}{\brickheight}{\brickwidth}{\mmcolor}{\scalebox{\legotextscaling}{OE}}{\studheight}


    \foreach \repry in {{\arrowsy+\brickwidth/2+\margin}, {\arrowsy-\brickwidth/2}, \bottombricky} {
        \foreach \posty in {{\arrowsy+\brickwidth/2+\margin}, {\arrowsy-\brickwidth/2}, \bottombricky} {
            \draw[arrow] (\reprx+\brickheight+\margin, \repry+\brickwidth/2) -- (\postx-\margin, \posty+\brickwidth/2);
        }
    }


    \node[grouplabel, align=center] (post_label) at (\postx+\brickheight/2, 4.0) {\scalebox{\textscaling}{Postprocessing}\\[-5pt]\scalebox{\textscaling}{}};
    \foreach \y/\label in {{\arrowsy+\brickwidth/2+\margin}/MSP, {\arrowsy-\brickwidth/2}/OpenMax} {
        \legoblock{\postx}{\y}{\brickheight}{\brickwidth}{\postcolor}{\scalebox{\legotextscaling}{\label}}{\studheight}
    }
    \node at ({\postx+(\brickheight-\studheight)/2}, 0.8) {\scalebox{1.1}{$\mathbf{\vdots}$}};
    \legoblock{\postx}{\bottombricky}{\brickheight}{\brickwidth}{\macolor}{\scalebox{\legotextscaling}{PostMax}}{\studheight}

    \draw[arrow] (\postx+\brickheight+\margin, \arrowsy) -- (\scoresx-\margin, \arrowsy);


    \newcommand{\barelement}[7]{
    \pgfmathsetmacro{\studheight}{#7}
    \pgfmathsetmacro{\studwidth}{#4/5}
    \pgfmathsetmacro{\studspacing}{(#4 - 2*\studwidth)/3}
    \pgfmathsetmacro{\firstStudY}{#2 + \studspacing}
    \pgfmathsetmacro{\secondStudY}{#2 + \studspacing + \studwidth + \studspacing}
    \pgfmathsetmacro{\centerX}{#1 + (#3-\studheight)/2}
    \pgfmathsetmacro{\centerY}{#2 + #4/2}
    \pgfmathsetmacro{\studX}{#1 + #3-\studheight}
    \pgfmathsetmacro{\highlightY}{#2 + #4/2}

    \draw[thick, fill=#5] (#1,#2) rectangle (#1+#3-\studheight,#2+#4);

    \draw[thick, fill=#5] (\studX,\firstStudY) rectangle (\studX+\studheight,\firstStudY+\studwidth);

    \draw[thick, fill=#5] (\studX,\secondStudY) rectangle (\studX+\studheight,\secondStudY+\studwidth);

    \node[align=center, text=black, rotate=270] at (\centerX,\centerY) {#6};
    }

    \def\scoresxcenter{\scoresx+2}


    \node[anchor=south west] (scores_small) at (\scoresx, 1.3) {\includegraphics[width=5.5cm]{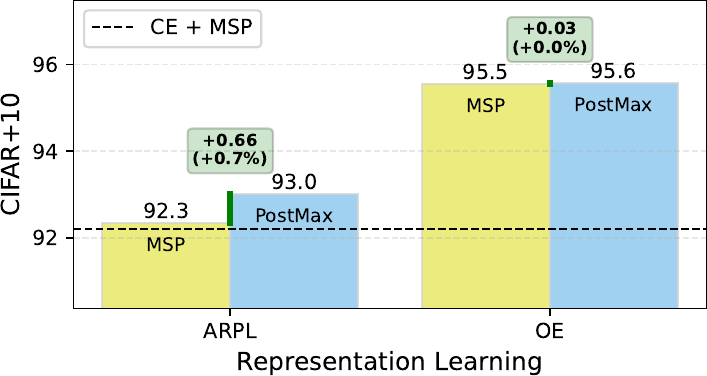}};
    \node[anchor=south west] (scores_small) at (\scoresx, -1.2) {\includegraphics[width=5.5cm]{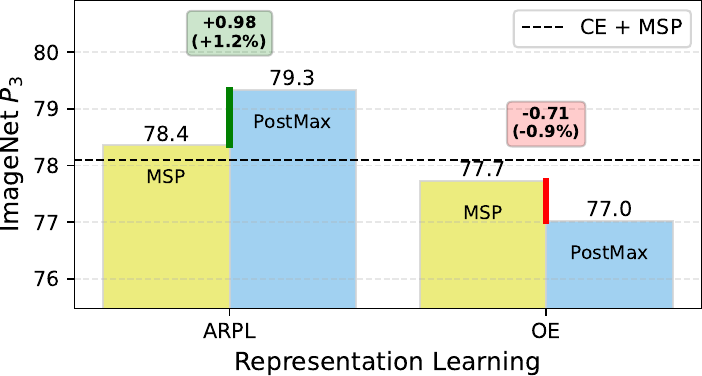}};





    \end{tikzpicture}
    }

    \Caption[fig:overview]{Open-Set Recognition Framework}{
        The modular two-stage OSR framework separates representation learning and postprocessing, and reveals additional performance gains, by leveraging interaction effects between the two.
        Undesirable interaction effects exist for magnitude-manipulating representation learning (red) combined with magnitude-aware postprocessors (blue), while beneficial effects are observed for \garbage{} classifiers and magnitude-aware postprocessors.
        Small-scale benchmarks are not predictive of large-scale performance and do not exhibit similar behavior due to limited similarity between known and known unknown classes.
    }

\end{figure*}

In both representation learning and postprocessing methods, the feature magnitude has been identified as a crucial factor for performance with unknown data observing smaller feature magnitudes \citep{dhamija2018agnostophobia,hendrycks2022scaling,vaze2022openset,cruz2024postmax,rabinowitz2025ghost}.
For instance, \citet{wang2025dissecting} highlight that \emph{magnitude-aware} (MA) OOD postprocessing generally outperforms alternatives. 
This raises the question: do MA postprocessors synergize with representation learning methods that purposefully manipulate feature magnitudes during training?
We identify a sub-category of representation learning methods, which we term \emph{magnitude-manipulating} (MM), that utilize known unknown data to separate known and known unknown classes during training based on feature magnitude.
However, MM methods tend to be sensitive to known unknown data distribution at large scale, with performance falling below baselines, despite their success on small-scale datasets \citep{hendrycks2021natural,wang2025dissecting}. 
This raises concerns about their real-world applicability, and we seek to understand the underlying causes of this performance degradation at scale.

Our analysis reveals the mechanism behind the performance degradation of MM methods as the interplay of magnitude-manipulation and high similarity between known unknown and known classes, a scenario not found on small-scale benchmarks.
This causes a \emph{magnitude collapse} in similar known classes and creates an undesirable linear dependency between feature magnitudes and class-wise accuracy, leading to systematically imbalanced class-wise detection performance. 
We show that magnitude collapse can be avoided by using an additional output node (a \garbage{}) for known unknown data, a simple and effective baseline that consistently outperforms other methods across scales and does not require hyperparameter tuning.
This performance degradation of MM methods is further amplified by MA postprocessing, which otherwise experiences desirable interaction effects when combined with non-MM representation learning methods and outperforms other postprocessors.
Moreover, our experiments suggest that representation learning without known unknown data and postprocessors are highly independent components of OSR systems with clear separation of responsibilities, enabling modular additive performance gains.

Overall, our study advances the understanding of OSR systems and provides actionable guidelines for researchers and practitioners.
First, leverage the modularity of OSR systems by augmenting non-MM representation learning methods with MA postprocessors to achieve additional performance gains almost for free.
Second, when training or fine-tuning a classifier with known unknown data that is visually similar to known classes, avoid MM representation learning methods.
Instead, use a \garbage{} classifier as baseline to mitigate magnitude collapse and leverage positive interaction effects.
Finally, validate OSR methods on large-scale benchmarks before deployment, as small-scale evaluations are not predictive of large-scale performance.

In summary, our contributions are as follows:
\begin{itemize}[leftmargin=1.5em]
  \item For the first time in OSR literature, we explore the modularity and interaction effects of representation learning and postprocessing methods through two-stage OSR, revealing where modular performance gains can be achieved and where to avoid negative synergies.
  \item We discover the magnitude collapse mechanism behind performance degradation at scale and how it impacts interaction effects.
  \item We demonstrate how interaction effects and known unknown data \emph{can} be leveraged at scale to achieve state-of-the-art performance regardless of known unknown data distribution with the simple \garbage{} baseline and two-stage processing of OSR techniques.
  \item Our experiments highlight that small-scale evaluations are not predictive of large-scale performance when using known unknown data.
\end{itemize}

\section{Related work}
\label{sec:related}

\paragraph{OSR and Relation to OOD Detection.}
OSR is formalized as the task of accurately classifying samples from known classes while rejecting samples from semantically unknown classes \citep{scheirer2013toward}, thereby detecting test-time semantic shift \citep{vaze2022openset}.
OOD detection is a broader task \citep{yang2024ood} that aims to detect general distribution shift, which can include semantic or covariate shifts \citep{yang2024ood,wang2025dissecting,hendrycks2021natural} and is often posed as a binary classification problem \citep{hendrycks2017baseline,liang2017odin,liu2020energy,huang2021gradnorm,sun2021react}.
As a result, OSR and OOD detection differ in their evaluation protocols \citep{vaze2022openset,wang2025dissecting}: OSR partitions a single dataset into known and unknown classes to mitigate covariate shifts \citep{neal2018counterfactual,palechor2023protocols}, while OOD detection typically uses different datasets for in-distribution (ID) and OOD classes \citep{hendrycks2017baseline}.
Despite these differences, it has been indicated that methods that perform well on one task tend to perform well on the other \citep{vaze2022openset,yang2024ood,wang2025dissecting}.

\paragraph{Known Unknown Data in OSR.}
Known unknown samples serve as a proxy for unknown classes during training and are distinct from known or ID classes \citep{scheirer2014probability,dhamija2018agnostophobia}.
Known unknown samples are also referred to as outlier images \citep{kong2021opengan}, natural adversarial examples \citep{hendrycks2021natural}, and negative samples \citep{palechor2023protocols}.
Real known unknown data is used for OOD detection \citep{hendrycks2019oe,liu2020energy} and OSR \citep{dhamija2018agnostophobia,palechor2023protocols}, dating back to the earliest approaches \citep{scheirer2014probability}. 
While a large attention in OSR research is paid to artificially generate known unknown samples \citep{ge2017generative,neal2018counterfactual,chen2020learning,chen2021arpl}, in this study we exclude generative methods and instead use real known unknown data.
The standard small-scale benchmarks MNIST, CIFAR, SVHN, and TinyImageNet partition all classes into known and unknown \citep{neal2018counterfactual}, therefore do not allow any known unknown classes.
The large-scale Semantic Shift Benchmark (SSB) \citep{vaze2022openset} uses the entire ImageNet-1K dataset as known classes and selects unknowns from a set of disjoint classes from ImageNet-21K-P.

\paragraph*{Differences and Similarities to Prior Art.}
\citet{wang2025dissecting} recently acknowledged the distinction between representation learning and postprocessors and the potential for combined approaches in the context of disentangling OSR and OOD methods and benchmarks without analyzing component interactions.
However, they focus primarily on OOD detection methods, with 10 out of 12 methods being postprocessing, leaving modern OSR methods and their interaction effects unexplored.


\section{Modular two-stage framework for OSR}

\begin{figure}[t]
  \centering



  \begin{tikzpicture}[scale=.53]
    \node[draw, rectangle, fill=gray!20, minimum width=5.3cm, minimum height=4cm, anchor=center] at (4.7,0) {};
    \node[anchor=north west] at (-.15,4.6) {\textbf{Representation Learning}};

    \node[draw, scale=.7](Image) at (-3,0) {\includegraphics[width=.2\textwidth]{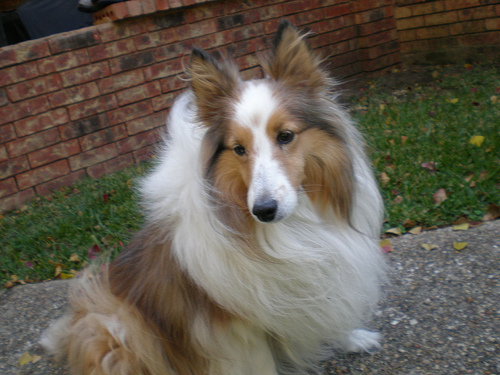}};
    \node[anchor=north ] at (-3,-1.3) {Input Image $\sample$};
    \node[draw,rectangle](Backbone) at (1.50,0) {
      \begin{turn}{270} Backbone Network $\backbone$\end{turn}
    };
    \node[draw,rectangle](Features) at (3.5,0) {\begin{turn}{270} Deep Features $\df$\end{turn}};
    \node[draw,rectangle](Logits) at (6,0) {\begin{turn}{270} Logits $\logits$\end{turn}};
    \node[draw,rectangle](Probabilities) at (8,0) {\begin{turn}{270} Probabilities $\confidence$\end{turn}};

    \node[draw, rectangle, fill=blue!20, minimum width=4.0cm, minimum height=4cm] at (13.7,0) {};
    \node[anchor=north west] at (10.2,4.6) {\textbf{Postprocessing}};

    \node[draw,rectangle](PostProcessor) at (11.5,0) {
      \begin{turn}{270} Postprocessor\end{turn}
    };
    \node[draw,rectangle,inner xsep=-3pt](ClassPred) at (14.5,2) {
      \begin{tabular}{c}
        Class \\
        Prediction $\pred$
      \end{tabular}
    };
    \node[draw,rectangle](Score) at (14.5,-2) {
      \begin{turn}{360} OOD Score $\score$\end{turn}
    };

    \draw[->,thick,dashed] (Image) -> (Backbone);
    \draw[->,thick] (Backbone) -> (Features);
    \draw[->,thick] (Features) -> node [above] {\begin{turn}{270} Linear \end{turn}} node [below]{\begin{turn}{270} $\logitW$\end{turn}} (Logits);
    \draw[->,thick,dashed] (Logits) -> node [above] {\begin{turn}{270} SoftMax\end{turn}} (Probabilities);

    \draw[->,thick,dashed] (Features) -- ++(0,3) -- ++({11.5-3.5},0) -> (PostProcessor);
    \draw[->,thick,dashed] (Logits) -- ++(0,-3) -- ++({11.5-6},0)  -> (PostProcessor);
    \draw[->,thick,dashed] (Probabilities) --  (PostProcessor);

    \draw[->,thick] (PostProcessor) -> (ClassPred);
    \draw[->,thick] (PostProcessor) -> (Score);
  \end{tikzpicture}

  \Caption[fig:framework]{Two-Stage Processing Framework for OSR}{
    An image is presented to the backbone network $\backbone$, which extracts deep features $\df$ that are then processed with a linear layer $\logitW$ to logits $\logits$, and further with SoftMax to probabilities $\confidence$.
    Solid lines indicate (potentially) learnable connections, while dashed lines highlight non-learnable connections.
    The postprocessor takes the deep features, logits, or probabilities as input and outputs a class prediction $\pred$ and a OOD score $\score$.
  }
\end{figure}

\begin{table*}[t]
  \centering

  \begin{subfigure}{.999\textwidth}
    \centering\footnotesize
    \begin{tabular}{|l|c|p{3em}<{\centering}|c|}
      \hline
      \textbf{Method} & \textbf{Known unknown} & \textbf{MM} & \textbf{Output $K+1$} \\
      \hline
      Cross-Entopy (CE) & none & & \\
      \hline
      ARPL \citep{chen2021arpl} & none & & \\
      \hline
      \garbage{} \citep{palechor2023protocols} & real & & Yes \\
      \hline
      Objectosphere (OS) \citep{dhamija2018agnostophobia} & real & Yes & \\
      \hline
      Outlier Exposure (OE) \citep{hendrycks2019oe} & real & Yes & \\
      \hline
    \end{tabular}
    \caption{Representation Learning}
    \label{tab:learning_methods}
  \end{subfigure}\\[.5em]
  \begin{subfigure}{.999\textwidth}
    \centering\footnotesize
    \begin{tabular}{|l|c|p{3em}<{\centering}|c|}
      \hline
      \textbf{Method} & \textbf{Trainable} & \textbf{MA} & \textbf{Inputs} \\
      \hline
      Maximum Softmax (MSP) \citep{hendrycks2017baseline} & & & $\confidence$ \\
      \hline
      MaxLogits/MLS \citep{hendrycks2022scaling,vaze2022openset} & & Yes & $\logits$ \\
      \hline
      OpenMax \citep{bendale2016openmax} & Yes &  & $\df$ \\
      \hline
      PostMax \citep{cruz2024postmax} & Yes & Yes & $\df,~\logits$ \\
      \hline
      GHOST \citep{rabinowitz2025ghost} & Yes & Yes & $\df,~\logits$ \\
      \hline
    \end{tabular}
    \caption{Postprocessing}
    \label{tab:postprocessing_methods}
  \end{subfigure}
  \Caption[tab:methods]{Key Characteristics}{
    We list key characteristics of \subref{tab:learning_methods} representation learning \rl{} and \subref{tab:postprocessing_methods} postprocessing \pp{} methods for OSR.
    Representation learning methods highlight which types of known unknown data they use, whether they are Magnitude-Manipulating (MM), and whether they use an additional output node $K+1$ for the unknown class.
    For postprocessors, we list whether they require training, are Magnitude-Aware (MA), and which types of inputs they operate on.
    }
\end{table*}

We disentangle OSR methods into two modular sequential components: representation learning \rl{} (classifier training) and postprocessing \pp{} (of pre-computed representations).
Within this two-stage framework, every OSR system can be viewed as a combination of one representation learning method $\rl$ and one postprocessor $\pp$, denoted as \comb{}{}, \eg, our baseline \comb{CE}{MSP} combines Cross-Entropy (CE) training with Maximum SoftMax Probability (MSP) postprocessing.
\fig{framework} provides an overview of the two-stage framework.
We summarize key characteristics relevant to this study of representation learning and postprocessing methods in \tab{methods}.
\apx{method-selection} discusses how methods from \tab{methods} can be formalized in this framework.

\paragraph*{Representation learning methods}
train or fine-tune a classifier \rl{} and extract the representations $\representation_n = (\df_n,\logits_n,\confidence_n) = \rl{}(\sample_n)$ for sample $\sample_n$, where $\df_n$ are discriminative deep features, $\logits_n$ are the logits, and $\confidence_n$ the probability distribution over the known classes.
They methods can modify the training process in various ways, typically by adapting the loss function \citep{dhamija2018agnostophobia,hendrycks2019oe,chen2020learning,chen2021arpl}, involving data augmentation, such as mixup \citep{zhang2018mixup,verma2019manifold} or generative methods \citep{ge2017generative,neal2018counterfactual,verma2019manifold,kong2021opengan,chen2021arpl,wilson2023safe,huang2023cssr}, or combinations thereof \citep{zhou2021placeholders}.
They are trained on a dataset $\knowns_\textrm{train} \cup \aux_\textrm{train}$, where for input $\sample_n$, $\knowns = \{(\sample_n, \clabel_n) \mid \clabel_n\in\knownlabels\}$ is the set of known samples with known class labels $\knownlabels=\{1,\ldots,K\}$, and $\aux = \{(\sample_n,\clabel_n) \mid \clabel_n \notin \knownlabels\}$ is the set of known unknown samples.
Evaluation uses $\knowns_\textrm{test} \cup \unknowns_\textrm{test}$, where $\unknowns = \{(\sample_n, \clabel_n)\mid \clabel_n \notin \knownlabels\}$ denotes unknown samples.
Note that while known unknown and unknown samples do not require a specific target label, they are required to not share the label space with known classes $\knownlabels$ \citep{scheirer2013toward}.

\paragraph*{Postprocessors}
operate post-hoc on representations $\representation_n$ to add open-set capabilities to a pre-trained closed-set classifier,\footnote{\label{fn:pp-class-prediction}Note that most postprocessors perform class predictions $\pred$ based on the argmax of the logits (or monotonic transformations thereof) and therefore yield identical class predictions and closed-set accuracy, addressing exclusively the separation between known and unknown classes.
} making them a cheap alternative to expensive representation learning training. 
However, postprocessors cannot undo any damage to the feature representation learned by the pre-trained network, \eg, when deep feature distributions $\df$ from known and unknown classes overlap, no postprocessor is able to separate those samples.
Postprocessors can involve training a secondary classifier \citep{scheirer2014probability,rudd2017evm}, employing a statistical model \citep{bendale2016openmax}, modifying the inputs \citep{liang2017odin}, or simply returning elements of $\representation_n$ \citep{hendrycks2017baseline,hendrycks2022scaling}.
We formalize postprocessors as follows: for test sample $\sample_n^{*}$ with $\representation_n^{*}$, we require a postprocessor to produce two outputs $\scores_n^{*} = (\pred_n, \score_n) = \pp{}(\representation_n^{*})$, where $\pred_n\in\knownlabels$ is the prediction of a known class label and $\score_n$ is an OOD score, where high $\score_n$ scores indicate known classes.
In an operational setting with a threshold $\threshold$, the OSR decision function can be defined as:

\begin{equation}
  \label{eq:osr-decision}
  \decision_\mathrm{OSR}(\representation_n^{*};\threshold) = \begin{cases}
    \pred_n & \text{if } \score_n \geq \threshold \\
    \mathrm{unknown} & \text{otherwise}
  \end{cases}
\end{equation}

\paragraph{Attributing performance contributions.}
To isolate and evaluate the performance contributions of individual components, we propose to measure the effect of representation learning or postprocessing by computing \emph{contribution deltas} of the respective components, for a given metric $\metric$.
We compute the \emph{representation learning contribution delta} to the CE baseline as a function of the postprocessor.
Similarly, we separate improvements of postprocessors by computing the \emph{postprocessor contribution delta} to the MSP baseline.
We define the representation learning and postprocessor contribution deltas as follows:

\begin{align}
  \label{eq:performance-delta-rl}
  \deltarl{i} &:= \metric(\comb{i}{}) - \metric(\comb{CE}{})\\
  \label{eq:performance-delta-pp}
  \deltapp{i} &:= \metric(\comb{}{i}) - \metric(\comb{}{MSP})
\end{align}
This allows us to decompose the performance improvement of any two-stage OSR system \comb{i}{j} over the \comb{CE}{MSP} baseline into representation learning and postprocessing contributions as follows:
\begin{align}
  \label{eq:perfrmance-decomposition-msp}
  \metric(\comb{i}{j}) - \metric(\comb{CE}{MSP}) &= \deltarl[MSP]{i} + \deltapp[i]{j} \\
  \label{eq:perfrmance-decomposition-ce}
  &= \deltarl[j]{i} + \deltapp[CE]{j}
\end{align}

\section{Study Design and Experimental Setup}
\label{sec:experimental-setup}
We choose five representation learning approaches, based on whether the method requires known unknown data and whether it (explicitly or implicitly) manipulates the feature magnitude.
As such we selected the following methods: Cross-Entropy (CE), ARPL \citep{chen2021arpl}, \garbage{} \citep{ren2019likelihood}, Outlier Exposure (OE) \citep{hendrycks2019oe}, and ObjectoSphere (OS) \citep{dhamija2018agnostophobia}.
We also choose five postprocessors, based on whether they take the feature magnitude into account, \ie, whether they are magnitude-aware.
We select MSP \citep{hendrycks2017baseline}, MaxLogits/MLS \citep{hendrycks2022scaling,vaze2022openset}, OpenMax \citep{bendale2016openmax}, PostMax \citep{cruz2024postmax} and GHOST \citep{rabinowitz2025ghost}.

\paragraph{Datasets.}
We conduct small-scale experiments on the standard OSR \textbf{CIFAR+N} benchmarks with $N\in\{10,50\}$ \citep{neal2018counterfactual}.
These protocols randomly sample 4 known classes from CIFAR-10 and N unknown classes from CIFAR-100 \cite{krizhevsky2009cifar}.\footnote{We use the same class allocations as \citet{chen2021arpl} for comparability.} 
To allow training with known unknown samples, we randomly sample N known unknown classes from the remaining classes in CIFAR-100.
Large-scale experiments are conducted on the \textbf{ImageNet} protocols \p1, \p2, and \p3 \citep{palechor2023protocols} based on the ILSVRC 2012 dataset \citep{russakovsky2015imagenet}.
These protocols offer increasing levels of difficulty from \p1 to \p3 by increasing semantic similarity between known, known unknown, and unknown classes based on the WordNet hierarchy \citep{miller1998wordnet}.
While \p1 poses an easy open-set task with low similarity between known and known unknown classes, \p2 and \p3 pose increasingly difficult open-set tasks with high similarity between known and known unknown classes.

\paragraph{Evaluation Metrics.}
To evaluate the binary unknown rejection and the closed-set performance in isolation, we use the Area Under the Receiver Operating Characteristics (AUROC) curve and closed-set accuracy, respectively.
To evaluate OSR performance we use Correct Classification Rate (CCR) \citep{dhamija2018agnostophobia} and False Positive Rate (FPR) and their single-valued derivations: Area Under the Open-Set Classification Rate (AUOSCR) curve \citep{vaze2022openset}, which provides a threshold-agnostic measure, and the Operational Open-set Accuracy (OOSA) \citep{cruz2024postmax}, which equally weights known and unknown samples and measures performance at an operational threshold.
For detailed descriptions of all metrics, please refer to \apx{performance-metrics}.
Given knowns $\knowns$, unknowns $\unknowns$, predictions $\scores_n^{*} = (\pred_n, \score_n)$, and threshold $\threshold$ we compute CCR and FPR following \citet{rabinowitz2025ghost} to allow for arbitrary OOD scores $\score$:

\begin{align}
  \label{eq:oscr}
  \mathrm{CCR}(\threshold) &= \frac{\bigl|\{(\sample_n, \clabel_n) \in \knowns \wedge \pred_n = \clabel_n \wedge \score_n \geq \threshold \}\bigr|}{\mid \knowns \mid}\\
  \mathrm{FPR}(\threshold) &= \frac{\bigl|\{(\sample_n, \clabel_n) \in \unknowns \wedge \score_n \geq \threshold \}\bigr|}{\mid \unknowns \mid}
\end{align}

To evaluate our methods we use multiple metrics to capture different aspects of performance.
AUOSCR and OOSA lack the interpretability of closed-set accuracy and AUROC, and therefore should not be solely relied upon.
On the other hand, the latter do not measure OSR performance holistically \citep{wang2022openauc}.
For example, closed-set accuracy is incapable of measuring interaction effects because all postprocessors except OpenMax exclusively rely on the logit order from upstream representation learning methods to produce predictions $\pred$ and ignore $\score$, resulting in identical accuracy scores (\sfig{supplemental-accuracy-heatmaps} in the Supplemental).
AUROC measures the ability to separate known and unknown samples across all thresholds via $\score$ and ignores $\pred$, thereby directly measuring the discriminative ability of the postprocessor and indirectly of the representation learning method.
AUOSCR and OOSA however jointly measure the quality of $\pred$ and $\score$ for any given threshold $\threshold$ via $\mathrm{CCR}(\threshold)$ and $\mathrm{FPR}(\threshold)$.
As such we rely primarily on AUOSCR and OOSA to answer our research questions but draw complementary insights from closed-set accuracy and AUROC.


\paragraph{Training Details.}
We train all networks from scratch, both to prevent leakage from pre-training on unknown classes and to isolate representation learning from fine-tuning effects.
All networks are trained using SGD with momentum of 0.9 and an initial learning rate of 0.1 with cosine annealing schedule \citep{loshchilov2017sgdr}.
For large-scale experiments we train ResNet50 for 120 epochs with batch size of 32 and weight decay of 1$e$-4.
The small-scale experiments are trained with the CNN architecture from \citet{neal2018counterfactual} and \citet{chen2021arpl} for 100 epochs with batch size of 128 and weight decay of 5$e$-4.
We perform early stopping according to validation confidence \citep{palechor2023protocols}.
For CE and ARPL, we compute validation confidence on known classes only since including known unknowns yielded unreliable results.
Where possible, we rely on recommended hyperparameters for each representation learning and postprocessing method.
Others are optimized on the validation set via grid search.
Optimal hyperparameters and their ranges are reported in \apx{hyperparameter-optimization}.

\section{Results and Discussion}
\label{sec:results}

\subsection{Representation Learning with Known Unknown Data}
\label{sec:scaling}

The first set of experiments aims at answering if representation learning with known unknown data can improve OSR performance on large-scale datasets, despite recent studies that suggest otherwise \citep{wang2025dissecting}.
We compare the performance of representation learning methods with known unknown data (\garbage{}, OE, and OS) to methods that only utilize known classes (CE and ARPL) across datasets.
Here, we ignore interaction effects and consider MSP postprocessing or aggregate results over all postprocessors.
The AUOSCR and OOSA for every \comb{}{} combination are shown in \fig{auoscr-heatmaps}, other metrics are reported in Tables~\ref{tab:summary-table-small} and \ref{tab:summary-table-large} in the Supplemental.

\begin{figure*}[t]
  \centering
  \includegraphics[width=\textwidth]{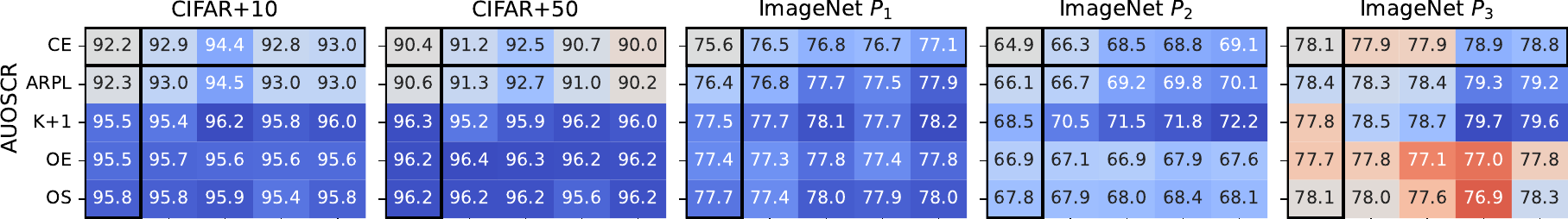}\\[0.2em]
  \includegraphics[width=\textwidth]{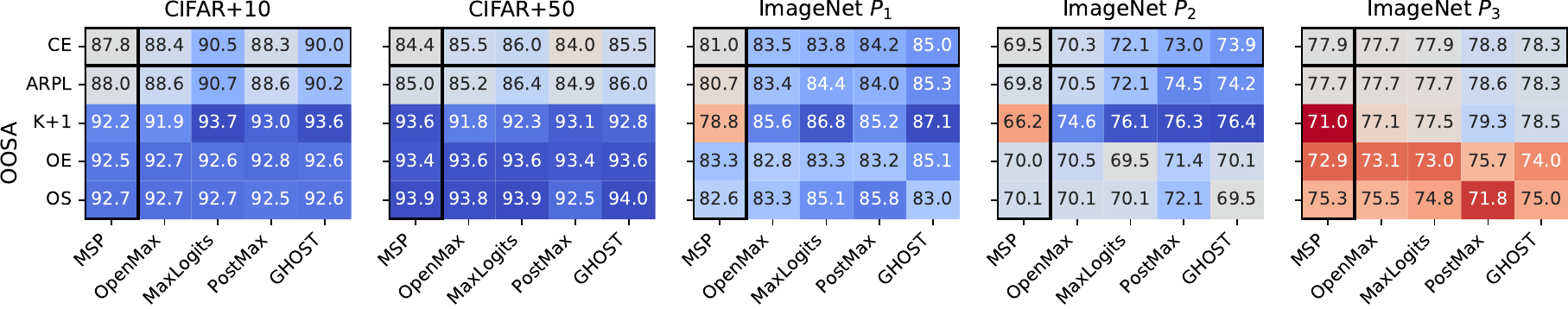}
  \Caption[fig:auoscr-heatmaps]{OSR Performance}{
    We show OSR performance for \comb{\rl}{\pp} combinations across datasets in AUOSCR (top) and OOSA (bottom).
    The color of each heatmap is normalized independently and centered at the \comb{CE}{MSP} baseline, where blue shows an increase and red a decrease. Single-stage methods are surrounded by a black border.
    Results for CIFAR+N are averaged over 5 trials.
  }
\end{figure*}

\paragraph{Small-Scale Training Excels on Known Unknown Data.}
On CIFAR+N, all representation learning methods utilizing known unknown data systematically outperform those that do not by up to 5.9 percentage points in AUOSCR with MSP. 
State-of-the-art ARPL achieves consistent but negligible improvements over CE for any given postprocessor.
With known classes being held constant between CIFAR+10 and CIFAR+50, we can see that additional known unknown data consistently improves AUOSCR by over 2 percentage points, even when evaluated on more unknown classes.
For all representation learning methods, the variations across postprocessors are comparably small, suggesting that representation learning contributes more toward overall performance than postprocessors on small-scale benchmarks, likely due to the limited diversity between known classes and the large distributional gap between known and unknown classes that can easily be exploited by representation learning methods with known unknown data.
Only MaxLogits provides substantial gains, up to 2.2 percentage points of AUOSCR over MSP for CE.

\iffalse
\begin{SCfigure}[1][t]
  \includegraphics[width=.59\columnwidth]{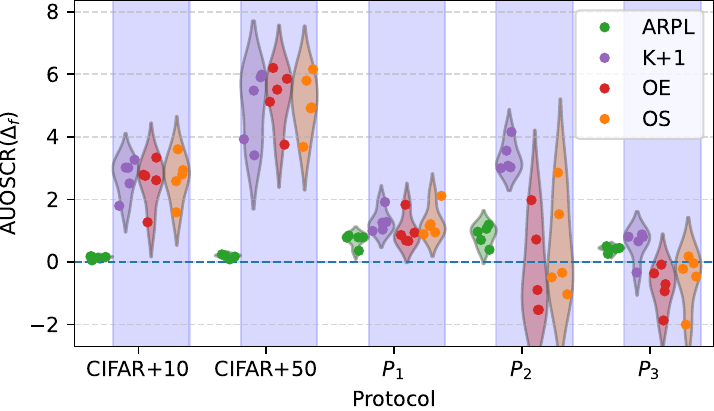}
  \caption{The representation learning performance contribution \deltarl in AUOSCR is shown as distributions over postprocessors and across protocols, with \p1, \p2, and \p3 increasing in similarity between known and known unknown classes. Methods that use known unknown data are marked with blue background. CIFAR+N results are averaged over 5 trials.}
  \label{fig:deltas-training}
\end{SCfigure}
\else
\begin{figure}[t]
  \centering
  \includegraphics[width=.7\columnwidth]{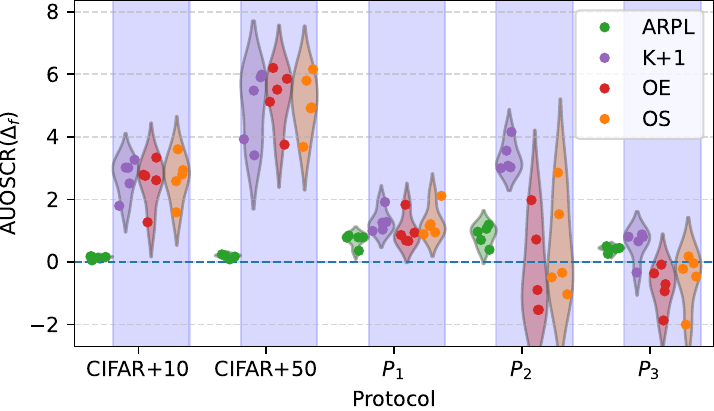}
  \Caption[fig:deltas-training]{Aggregated Results}{
    The representation learning performance contribution \deltarl{} in AUOSCR is shown as distributions over postprocessors and across protocols. Methods that use known unknown data are marked with blue background. CIFAR+N results are averaged over 5 trials.
    }
\end{figure}
\fi

\paragraph{Performance Degradation at Large Scale.}
On large-scale ImageNet protocols, this outperformance from using known unknown data vanishes, even underperforming the \comb{CE}{MSP} baseline on \p3 for most postprocessors, in line with \citet{wang2025dissecting}.
\fig{deltas-training} depicts the representation learning contribution deltas $\deltarl[j]{i}$ for each representation learning method $i$ across datasets as distribution over all postprocessors $j$.
The representation learning contribution delta for MM methods OE and OS degrades and turns negative with increasing similarity of known and known unknown classes on \p2 and \p3, destroying performance across most postprocessors.
\citet{wang2025dissecting} attribute their findings of poor OE performance to low correlation between known unknown and unknown classes or high correlation between known and unknown classes. 
However, the \garbage{} classifier does not experience this performance degradation with identical data, demonstrating that it cannot be explained by the data distributions alone but by the interplay between data distributions and training objective.
Strong OOD detection performance with MSP (see AUROC in \sfig{supplemental-auroc-heatmaps}) and high correlation between AUOSCR and closed-set accuracy (cf.~\sfig{supplemental-accuracy-heatmaps} in Supplemental) suggest that the performance degradation can partially be explained by a loss of discriminative information for known classes.


\begin{figure}[t]
  \centering
  \includegraphics[width=.95\columnwidth]{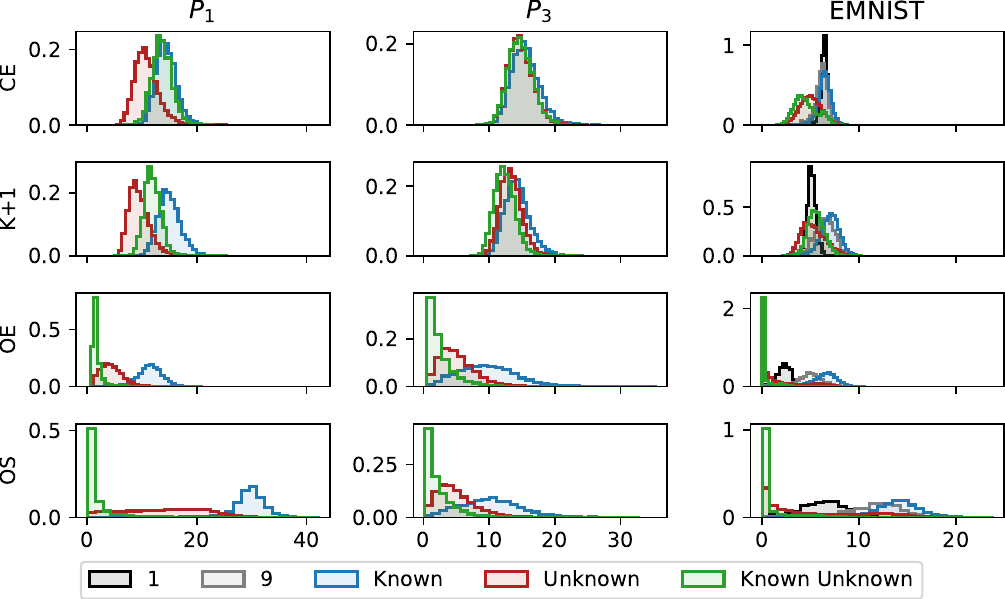}
  \Caption[fig:magnitudes]{Feature Magnitudes}{
    We plot feature magnitude distributions of known, known unknown, and unknown classes on protocols \p1, \p3, and EMNIST.
    OE and OS experience feature magnitude collapse on \p3 and EMNIST, pulling their feature magnitudes towards zero.
    For EMNIST, we show distributions for known classes 1 and 9 (black and grey) that are highly similar to known unknown classes, and other known classes (blue) separately.
  }
\end{figure}

\paragraph{Magnitude Collapse.}
To understand \emph{why} MM methods degrade in accuracy, AUOSCR, and OOSA, we analyze the feature magnitude distributions of known, known unknown, and unknown samples (\fig{magnitudes}).
With highly similar known unknown samples, MM methods inadvertently draw features of known classes towards the origin of the feature space, resulting in \emph{magnitude collapse} and effectively overlapping them with known unknown, unknown, and other known samples.
We analyze the relationship between feature magnitude and class-wise CCR \citep{rabinowitz2025ghost} at the operational threshold, $\mathrm{CCR}_c(\threshold^\star)$, see \apx{performance-metrics}, and perform linear regression\footnote{All regression assumptions are reasonably met.} against the class-wise average feature magnitude for each known class $c$ (\fig{supplemental-regressions} in the Supplemental).
On CIFAR+N, only MM methods exhibit a statistically significant positive correlation between average feature magnitude and $\mathrm{CCR}_c(\threshold^\star)$, with large effect sizes of $R^2$ up to 80\%.
On the easily separable \p1, most models exhibit significant positive correlations, but often with meaningless effect sizes below 10\%.
However, with increasing similarity in known and known unknown samples on \p2 and \p3, MM methods learn much stronger and statistically significant relationships, resulting in practically significant effect sizes of $R^2$ up to 38\% on \p3 (\fig{regression-p3}).
This shows how magnitude-manipulation increases the dependency between feature magnitude and CCR by systematically reducing performance on a few classes in a trade-off to maintain \emph{overall} high binary ID-vs-OOD separation via MSP (see AUROC \sfig{supplemental-auroc-heatmaps}) and increasing the minimal class-wise CCR.

In contrast, \garbage{} counteracts magnitude collapse by learning sufficiently large feature magnitudes to achieve high SoftMax probabilities for the additional output node during training.
Since \garbage{} is trained via CE loss, it forces probabilities corresponding to any output node (known or known unknown) to be close to 1 for samples of the respective class.
Formally, we want the trained network to be able to achieve $\confidence_k > 1-\epsilon$ for any sample and output class $c\in\{1,\ldots,K\}$, for some small $\epsilon>0$.
This holds if (but not only if) the difference between the maximal two logits surpasses a lower bound $\lb$, which is equivalent to (see \apx{derivation_feature_magnitude}):
\begin{equation}
  \label{eq:sufficient-condition}
  {\|\df\|}_2 > \frac{\lb}{\delta_{\logits_c/ {\|\df\|}_2}} = \frac{\log(\frac{1-\epsilon}{\epsilon}) + \log(K-1)}{{\|\logitW_c\|}_2 \cos(\dfangle_c) - \max_{c^\prime\neq c}\{{\|\logitW_{c^\prime}\|}_2 \cos(\dfangle_{c^\prime})\}} \Longrightarrow \confidence_c > 1-\epsilon \,,
\end{equation}
where $\delta_{\logits_c {\|\df\|}_2}$ is the difference logits divided by their feature magnitude.
Note that $\lb$ is constant for any dataset, loss and $\epsilon$, and $\delta_{\logits_c/ {\|\df\|}_2}$ is generally small because weight magnitudes are minimized through weight decay and the cosines are bounded between -1 and 1.
This provides upward pressure on feature magnitudes during training and requires deep feature magnitudes to exceed a lower bound for any sample given a trained classifier.
In other words, learning sufficiently large feature magnitudes is a sufficient condition to achieve high confidence for some class given trained classifier weights $\logitW$.
This training incentive prevents magnitude collapse in \garbage{} and ensures that \garbage{} learns feature magnitude distributions similar to CE and ARPL.
Surprisingly, despite training on known unknown data, \garbage{} learns lower feature magnitudes for known unknown and unknown classes while maintaining sufficiently large feature magnitudes to achieve high confidences for known and known unknown classes.
The difference in feature magnitudes between known, known unknown, and unknown classes is analogous to CE but slightly more pronounced on $\p3$, see
\fig{magnitudes-complete}.
This provides theoretical support for combinations of \garbage{} with MA postprocessors, because high level magnitude distribution statistics are similar to CE, for which those postprocessors were designed.

\iffalse
\begin{SCfigure}[1][t]
  \centering
  \includegraphics[width=.70\columnwidth]{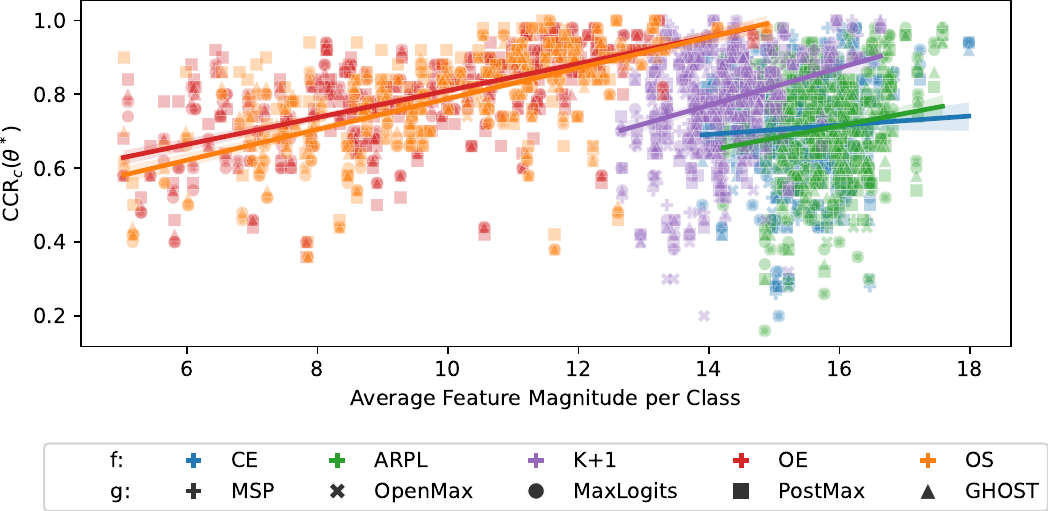}
  \caption{
    Linear relationship between class-wise CCR, $\mathrm{CCR}_c(\threshold^\star)$, at the operational threshold and class-wise average feature magnitude for known classes on \p3.
    We perform regression for each representation learning method independently and over all postprocessors.
  }
  \label{fig:regression-p3}
\end{SCfigure}
\else
\begin{figure}[t]
  \centering
  \includegraphics[width=.80\columnwidth]{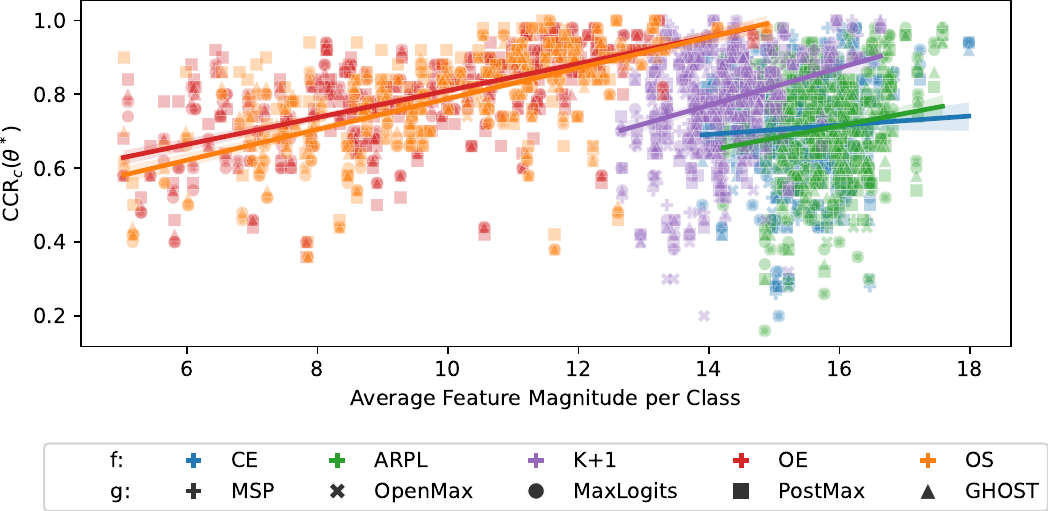}
  \Caption[fig:regression-p3]{CCR vs. Magnitude}{
    We exhibit a linear relationship between class-wise CCR, $\mathrm{CCR}_c(\threshold^\star)$, at the operational threshold and class-wise average feature magnitude for known classes on \p3.
    We perform regression for each representation learning method independently and over all postprocessors.
  }
\end{figure}
\fi

\paragraph{Qualitative Analysis on Class Similarity.}
We replicate and qualitatively investigate the feature collapse for MM methods at small scale by curating a hard EMNIST benchmark that includes known unknown classes which are highly similar to known class 1 and often indistinguishable to class 9 without context (see \apx{emnist} for details).
Experiments are identical to CIFAR+N but networks are trained for 50 epochs.
Observable from \fig{magnitudes} (right), for MM methods, the feature representations of digits 1 and 9 are pulled towards the origin.
We compute $\mathrm{CCR}_c(\threshold^\star)$ per representation learning method for classes 1, 9, and rest (all other known classes).
For OS and OE combined, these classes experience $\mathrm{CCR}_c(\threshold^\star)$ in the range [58.1\%, 88.7\%] (digit 1) and [93.6\%, 97.2\%] (digit 9), while other digits achieve over 96.7\% CCR, showing dramatically imbalanced detection performance on known classes.
See \tab{emnist-summary-table} in the Supplemental for all representation learning methods.
Non-MM methods have $\mathrm{CCR}_k(\threshold^\star)$ evenly distributed between 83.8\% and 98\% over all postprocessors.
This demonstrates how high similarity between known and known unknown classes, combined with magnitude-manipulation, irreparably damages feature representations and ruins classification of visually similar known classes.


\subsection{Interaction Effects}
\label{sec:interaction}

\begin{figure*}[t]
  \centering
  \begin{subfigure}{.99\textwidth}
    \includegraphics[width=.99\textwidth, clip, trim=0 20 0 0]{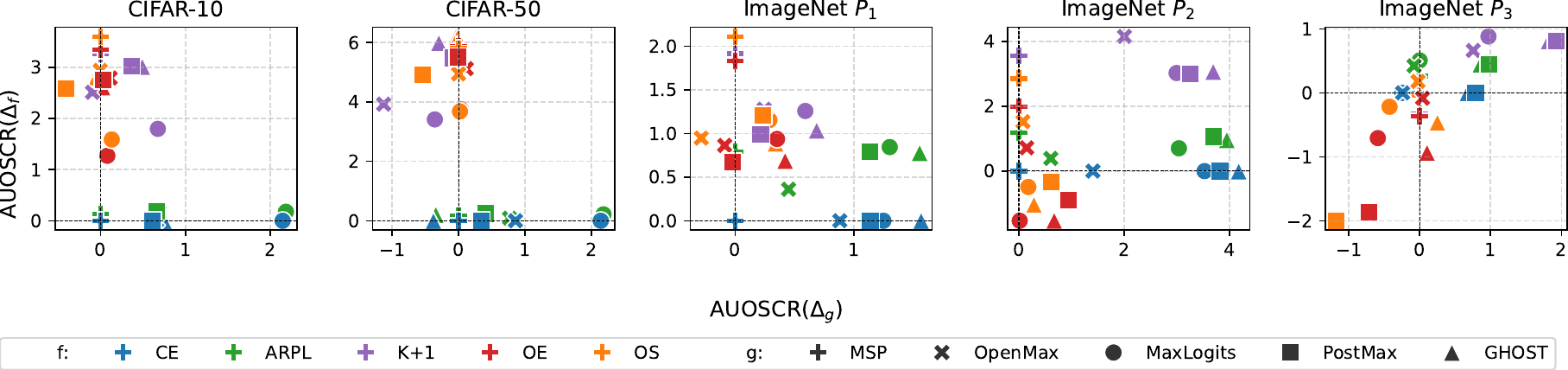}
  \end{subfigure}
  \begin{subfigure}{.99\textwidth}
    \includegraphics[width=.99\textwidth]{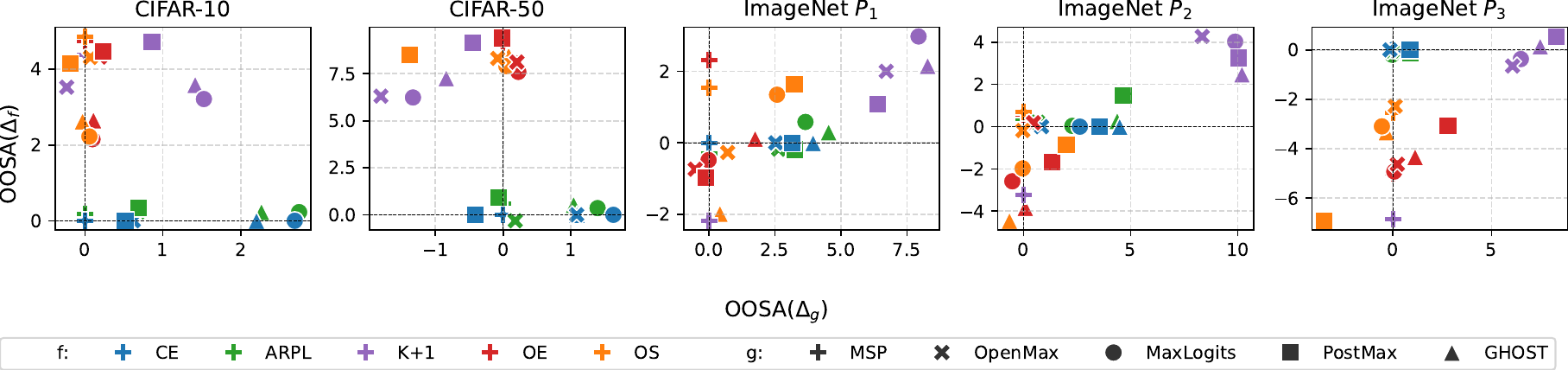}
  \end{subfigure}
  \Caption{Interaction effects}{We indicate interaction effects of representation learning and postprocessing components as correlation between representation learning contribution \deltarl{} and postprocessor contribution \deltapp{} for AUOSCR (top row) and OOSA (bottom row).}
  \label{fig:delta-scatter}
\end{figure*}

The second set of experiments aims to answer whether the optimal postprocessor should be informed by the representation learning method, and whether magnitude-manipulating representation learning can enhance the performance of magnitude-aware postprocessors.
We analyze the relationship and interactions between representation learning and postprocessor contributions by plotting the representation learning contribution delta \deltarl{} against the postprocessing contribution delta \deltapp{} for each combination \comb{}{} in \fig{delta-scatter}.

\paragraph{Independent Contributions Without Known Unknown Data.}
Across all evaluation metrics and protocols, all experiments reveal that postprocessor contributions are almost perfectly independent from representation learning contributions, when trained \emph{without} known unknown data, \eg, the postprocessor contribution of GHOST is independent of the used representation learning method: $\deltapp[CE]{GHOST} \approx \deltapp[ARPL]{GHOST} = 1.5$.
This suggests that OSR system components have separate responsibilities when trained on known classes only, with representation learning addressing the ID classification and postprocessors addressing the open-set capabilities.\footref{fn:pp-class-prediction}
This allows to freely combine representation learning and postprocessor, with magnitude-aware postprocessors consistently favored, supporting \citet{wang2025dissecting}, to achieve performance gains without risking undesirable interactions.

\paragraph{Interaction Effects With Known Unknown Data.}
Representation learning methods that train \emph{with} known unknown data only show interaction effects with high similarity of known and known unknown classes.
Interaction effects are characterized by a correlation between \deltarl{} and \deltapp{} in \fig{delta-scatter}.
On ImageNet benchmarks, representation learning methods with known unknown data show interaction effects, with particularly MA postprocessors strongly amplifying positive and negative gains from representation learning.
While magnitude-aware postprocessors consistently outperform others for non-MM methods, they can amplify performance degradation for MM methods due to their sensitivity to the magnitude collapse.
In particular, PostMax, and to a lesser extent GHOST, demonstrate both the highest gains for non-MM methods (8.3 percentage points or +11.7\% for \garbage{}), as well as the most severe performance degradation for MM methods (-1.2 percentage points or -1.5\% for OS) on \p3.
These interaction effects are even clearer when evaluated via OOSA, which equally weights known and unknown test samples.

\subsection{Robustness to Backbone Architecture}
To ensure that our findings are not specific to the ResNet architecture, we replicate the experiments on \p2 using a Swin-B transformer \citep{liu2021swin}.
Protocol \p2 balances difficulty and computational cost and provides a good trade-off between the two \citep{palechor2023protocols}, allowing us to evaluate the generalizability of our findings to a transformer-based architecture.
Training is conducted with identical hyperparameters to the ResNet-50 experiments, except for a reduced learning rate of $2e-4$ to account for a lower batch size of 90.
We trained 300 epochs with the AdamW optimizer and weight decay of 0.05 to ensure convergence.

Overall we observe similar relative performance as with ResNet-50 (\fig{swin}).
Particularly, the magnitude-manipulating methods OE and OS experience performance degradation to the \comb{CE}{MSP} baseline when using MSP postprocessing, and combined with magnitude-aware postprocessors they experience further degradation or mixed postprocessing contributions when evaluated via OOSA.
For AUOSCR and AUROC, the performance degradation is less severe.
As expected, \garbage{} systematically outperforms all other combinations when combined with magnitude-aware postprocessors across all metrics.
The absolute performance of all methods is lower compared to ResNet-50 backbones, likely due to the increased data requirement of a transformer coupled with a larger model and limited data set sizes due to the protocols containing only a fraction of the ImageNet dataset classes.
Interestingly, the training of the reciprocal points (ARPL) proved to be unstable under the AdamW optimizer, while the backbone training remained stable, resulting in significantly lower accuracy and consequently lower AUOSCR, OOSA, and AUROC.
While future work is needed to to investigate better training methods for ARPL with transformers, we observe that magnitude-aware postprocessors nontheless can recover most of the performance gap to CE for binary ID-vs-OOD separation, as shown by the AUROC results, highlighting the practical importance and utility of a two-stage approach to OSR by combining representation learning with postprocessing.

\begin{figure}
  \centering
  \includegraphics[width=\columnwidth]{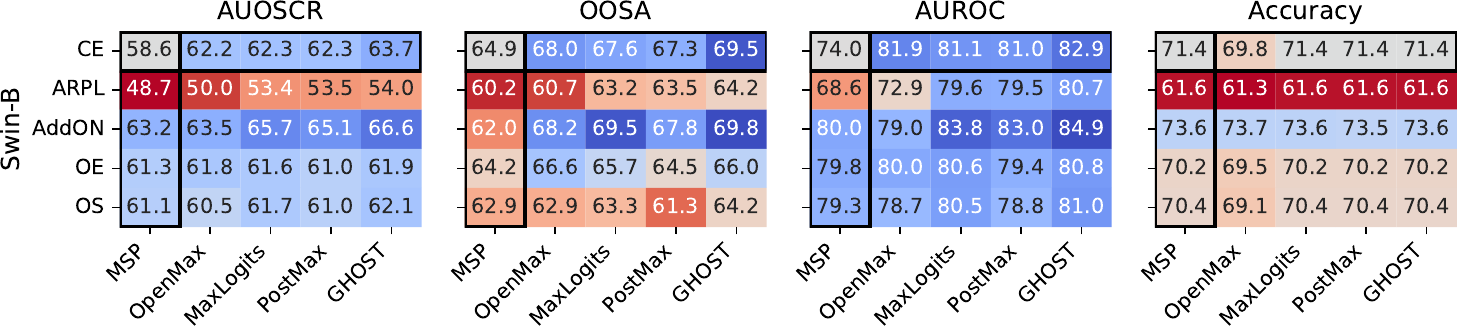}
  \Caption[fig:swin]{Swin Backbone}{We show OSR performance on \p2 with a Swin-B backbone, showing accuracy, AUROC, and AUOSCR metrics. Similar interaction effects between representation learning and postprocessing methods are observed as with ResNet-50 backbones.
  }
\end{figure}
\section{Conclusion}
\label{sec:conclusion}

In this study, we adopt a two-stage framework for modularly combining representation learning and postprocessing methods for Open-Set Recognition (OSR). 
We show that representation learning \emph{without known unknown data} leads to independent OSR components, that can be freely combined to achieve additive performance gains, whereas representation learning \emph{with known unknown data} can experience interaction effects with postprocessors that can improve or degrade OSR performance.
We explain this performance degradation and the key role of feature magnitude in interaction effects via the magnitude collapse mechanism, revealing several insights.
First, the similarity between known unknown and known classes is a key factor for performance degradation at scale, a scenario that does not occur on small-scale benchmarks.
Second, magnitude collapse creates an undesirable linear dependency between feature magnitude and class-wise detection performance on the in-distribution and OSR task, leading to systematically imbalanced detection across known classes.
Furthermore, we demonstrate how magnitude collapse can be circumvented by the simple yet effective baseline \garbage{} and that, contrary to recent findings, training with known unknown data \emph{can} improve OSR performance at any scale and regardless of known unknown data selection.
Our findings invalidate current best practices in OSR, demonstrating that small-scale evaluations with known unknown data do not translate to large-scale performance.
Generally, small-scale experiments are primarily driven by representation learning via inclusion of known unknown data and favor the simpler postprocessors like MaxLogits, while large-scale experiments benefit from combining representation learning and postprocessing, mainly through joining \garbage{} with MA postprocessors like PostMax or GHOST.
As such, representation learning methods considered state-of-the-art based on CIFAR benchmarks, such as Outlier Exposure, can degrade significantly below baselines at scale.


\section*{Acknowledgements}
We thank the University of Zurich and the Digital Society Initiative for financing this research through a research scholarship. 
Furthermore, we thank the University of Zurich for providing the computational resources.

\bibliography{References,Publications,Theses}

\begin{thebibliography}{39}
\providecommand{\natexlab}[1]{#1}
\providecommand{\url}[1]{\texttt{#1}}
\expandafter\ifx\csname urlstyle\endcsname\relax
  \providecommand{\doi}[1]{doi: #1}\else
  \providecommand{\doi}{doi: \begingroup \urlstyle{rm}\Url}\fi

\bibitem[Bendale and Boult(2016)]{bendale2016openmax}
Abhijit Bendale and Terrance~E. Boult.
\newblock Towards open set deep networks.
\newblock In \emph{Conference on Computer Vision and Pattern Recognition (CVPR)}. IEEE, 2016.

\bibitem[Chen et~al.(2020)Chen, Qiao, Shi, Peng, Li, Huang, Pu, and Tian]{chen2020learning}
Guangyao Chen, Limeng Qiao, Yemin Shi, Peixi Peng, Jia Li, Tiejun Huang, Shiliang Pu, and Yonghong Tian.
\newblock Learning open set network with discriminative reciprocal points.
\newblock In \emph{European Conference on Computer Vision (ECCV)}. Springer, 2020.

\bibitem[Chen et~al.(2021)Chen, Peng, Wang, and Tian]{chen2021arpl}
Guangyao Chen, Peixi Peng, Xiangqian Wang, and Yonghong Tian.
\newblock Adversarial reciprocal points learning for open set recognition.
\newblock \emph{Transactions on Pattern Analysis and Machine Intelligence (TPAMI)}, 44\penalty0 (11), 2021.

\bibitem[Cohen et~al.(2017)Cohen, Afshar, Tapson, and van Schaik]{cohen2017emnist}
Gregory Cohen, Saeed Afshar, Jonathan Tapson, and André van Schaik.
\newblock {EMNIST}: Extending {MNIST} to handwritten letters.
\newblock In \emph{International Joint Conference on Neural Networks (IJCNN)}, 2017.

\bibitem[Cruz et~al.(2024)Cruz, Rabinowitz, Günther, and Boult]{cruz2024postmax}
Steve Cruz, Ryan Rabinowitz, Manuel Günther, and Terrance~E. Boult.
\newblock Operational open-set recognition and postmax refinement.
\newblock In \emph{European Conference on Computer Vision (ECCV)}, 2024.

\bibitem[Dhamija et~al.(2018)Dhamija, G\"unther, and Boult]{dhamija2018agnostophobia}
Akshay~Raj Dhamija, Manuel G\"unther, and Terrance~E. Boult.
\newblock Reducing network agnostophobia.
\newblock In \emph{Advances in Neural Information Processing Systems (NeurIPS)}, 2018.

\bibitem[Dhamija et~al.(2020)Dhamija, G\"unther, Ventura, and Boult]{dhamija2020elephant}
Akshay~Raj Dhamija, Manuel G\"unther, Jonathan Ventura, and Terrance~E. Boult.
\newblock The overlooked elephant of object detection: Open set.
\newblock In \emph{Winter Conference on Applications of Computer Vision (WACV)}, 2020.

\bibitem[Djurisic et~al.(2023)Djurisic, Bozanic, Ashok, and Liu]{djurisic2023-ash}
Andrija Djurisic, Nebojsa Bozanic, Arjun Ashok, and Rosanne Liu.
\newblock Extremely {{Simple Activation Shaping}} for {{Out-of-Distribution Detection}}.
\newblock In \emph{The {{Eleventh International Conference}} on {{Learning Representations}}}, 2023.

\bibitem[Ge et~al.(2017)Ge, Demyanov, and Garnavi]{ge2017generative}
Zongyuan Ge, Sergey Demyanov, and Rahil Garnavi.
\newblock Generative {OpenMax} for multi-class open set classification.
\newblock In \emph{British Machine Vision Conference (BMVC)}, 2017.

\bibitem[Hendrycks and Gimpel(2017)]{hendrycks2017baseline}
Dan Hendrycks and Kevin Gimpel.
\newblock A baseline for detecting misclassified and out-of-distribution examples in neural networks.
\newblock In \emph{International Conference on Learning Representations (ICLR)}, 2017.

\bibitem[Hendrycks et~al.(2019)Hendrycks, Mazeika, and Dietterich]{hendrycks2019oe}
Dan Hendrycks, Mantas Mazeika, and Thomas Dietterich.
\newblock Deep anomaly detection with outlier exposure.
\newblock In \emph{International Conference on Learning Representations (ICLR)}, 2019.

\bibitem[Hendrycks et~al.(2021)Hendrycks, Zhao, Basart, Steinhardt, and Song]{hendrycks2021natural}
Dan Hendrycks, Kevin Zhao, Steven Basart, Jacob Steinhardt, and Dawn Song.
\newblock Natural adversarial examples.
\newblock In \emph{Conference on Computer Vision and Pattern Recognition (CVPR)}, 2021.

\bibitem[Hendrycks et~al.(2022)Hendrycks, Basart, Mazeika, Zou, Kwon, Mostajabi, Steinhardt, and Song]{hendrycks2022scaling}
Dan Hendrycks, Steven Basart, Mantas Mazeika, Andy Zou, Joseph Kwon, Mohammadreza Mostajabi, Jacob Steinhardt, and Dawn Song.
\newblock Scaling out-of-distribution detection for real-world settings.
\newblock In \emph{International Conference on Machine Learning (ICML)}. PMLR, 2022.

\bibitem[Huang et~al.(2023)Huang, Wang, Hu, and Cheng]{huang2023cssr}
Hongzhi Huang, Yu~Wang, Qinghua Hu, and Ming-Ming Cheng.
\newblock Class-specific semantic reconstruction for open set recognition.
\newblock \emph{Transactions on Pattern Analysis and Machine Intelligence (TPAMI)}, 45\penalty0 (4), 2023.

\bibitem[Huang et~al.(2021)Huang, Geng, and Li]{huang2021gradnorm}
Rui Huang, Andrew Geng, and Yixuan Li.
\newblock On the importance of gradients for detecting distributional shifts in the wild.
\newblock In \emph{Advances in Neural Information Processing Systems (NeurIPS)}, 2021.

\bibitem[Kong and Ramanan(2021)]{kong2021opengan}
Shu Kong and Deva Ramanan.
\newblock Opengan: Open-set recognition via open data generation.
\newblock In \emph{International Conference on Computer Vision (ICCV)}, 2021.

\bibitem[Krizhevsky and Hinton(2009)]{krizhevsky2009cifar}
Alex Krizhevsky and Geoffrey Hinton.
\newblock Learning multiple layers of features from tiny images.
\newblock Technical report, University of Toronto, 2009.

\bibitem[Liang et~al.(2017)Liang, Li, and Srikant]{liang2017odin}
Shiyu Liang, Yixuan Li, and Rayadurgam Srikant.
\newblock Enhancing the reliability of out-of-distribution image detection in neural networks.
\newblock In \emph{International Conference on Learning Representations (ICLR)}, 2017.

\bibitem[Liu et~al.(2020)Liu, Wang, Owens, and Li]{liu2020energy}
Weitang Liu, Xiaoyun Wang, John Owens, and Yixuan Li.
\newblock Energy-based out-of-distribution detection.
\newblock In \emph{Advances in Neural Information Processing Systems (NeurIPS)}, 2020.

\bibitem[Liu et~al.(2021)Liu, Lin, Cao, Hu, Wei, Zhang, Lin, and Guo]{liu2021swin}
Ze~Liu, Yutong Lin, Yue Cao, Han Hu, Yixuan Wei, Zheng Zhang, Stephen Lin, and Baining Guo.
\newblock Swin transformer: {Hierarchical} vision transformer using shifted windows.
\newblock In \emph{International Conference on Computer Vision (ICCV)}, 2021.

\bibitem[Loshchilov and Hutter(2017)]{loshchilov2017sgdr}
Ilya Loshchilov and Frank Hutter.
\newblock Sgdr: Stochastic gradient descent with warm restarts.
\newblock In \emph{International Conference on Learning Representations (ICLR)}, 2017.

\bibitem[Miller(1998)]{miller1998wordnet}
George~A Miller.
\newblock \emph{WordNet: An electronic lexical database}.
\newblock MIT press, 1998.

\bibitem[Neal et~al.(2018)Neal, Olson, Fern, Wong, and Li]{neal2018counterfactual}
Lawrence Neal, Matthew Olson, Xiaoli Fern, Weng-Keen Wong, and Fuxin Li.
\newblock Open set learning with counterfactual images.
\newblock In \emph{European Conference on Computer Vision (ECCV)}, 2018.

\bibitem[Palechor et~al.(2023)Palechor, Bhoumik, and G\"unther]{palechor2023protocols}
Andres Palechor, Annesha Bhoumik, and Manuel G\"unther.
\newblock Large-scale open-set classification protocols for imagenet.
\newblock In \emph{Winter Conference on Applications of Computer Vision (WACV)}, 2023.

\bibitem[Rabinowitz et~al.(2025)Rabinowitz, Cruz, Günther, and Boult]{rabinowitz2025ghost}
Ryan Rabinowitz, Steve Cruz, Manuel Günther, and Terrance~E. Boult.
\newblock {GHOST}: Gaussian hypothesis open-set technique.
\newblock In \emph{AAAI Conference on Artificial Intelligence}, 2025.

\bibitem[Ren et~al.(2019)Ren, Liu, Fertig, Snoek, Poplin, Depristo, Dillon, and Lakshminarayanan]{ren2019likelihood}
Jie Ren, Peter~J Liu, Emily Fertig, Jasper Snoek, Ryan Poplin, Mark Depristo, Joshua Dillon, and Balaji Lakshminarayanan.
\newblock Likelihood ratios for out-of-distribution detection.
\newblock In \emph{Advances in Neural Information Processing Systems (NeurIPS)}, 2019.

\bibitem[Rudd et~al.(2017)Rudd, Jain, Scheirer, and Boult]{rudd2017evm}
Ethan~M. Rudd, Lalit~P. Jain, Walter~J. Scheirer, and Terrance~E. Boult.
\newblock The extreme value machine.
\newblock \emph{Transactions on Pattern Analysis and Machine Intelligence (TPAMI)}, 2017.

\bibitem[Russakovsky et~al.(2015)Russakovsky, Deng, Su, Krause, Satheesh, Ma, Huang, Karpathy, Khosla, Bernstein, et~al.]{russakovsky2015imagenet}
Olga Russakovsky, Jia Deng, Hao Su, Jonathan Krause, Sanjeev Satheesh, Sean Ma, Zhiheng Huang, Andrej Karpathy, Aditya Khosla, Michael Bernstein, et~al.
\newblock Imagenet large scale visual recognition challenge.
\newblock \emph{International Journal of Computer Vision (IJCV)}, 115\penalty0 (3), 2015.

\bibitem[Scheirer et~al.(2013)Scheirer, de~Rezende~Rocha, Sapkota, and Boult]{scheirer2013toward}
Walter~J. Scheirer, Anderson de~Rezende~Rocha, Archana Sapkota, and Terrance~E. Boult.
\newblock Towards open set recognition.
\newblock \emph{Transactions on Pattern Analysis and Machine Intelligence (TPAMI)}, 35\penalty0 (7), 2013.

\bibitem[Scheirer et~al.(2014)Scheirer, Jain, and Boult]{scheirer2014probability}
Walter~J. Scheirer, Lalit~P. Jain, and Terrance~E. Boult.
\newblock Probability models for open set recognition.
\newblock \emph{Transactions on Pattern Analysis and Machine Intelligence (TPAMI)}, 36\penalty0 (11), 2014.

\bibitem[Sun et~al.(2021)Sun, Guo, and Li]{sun2021react}
Yiyou Sun, Chuan Guo, and Yixuan Li.
\newblock {ReAct}: Out-of-distribution detection with rectified activations.
\newblock In \emph{Advances in Neural Information Processing Systems (NeurIPS)}, 2021.

\bibitem[Vaze et~al.(2022)Vaze, Han, Vedaldi, and Zissermann]{vaze2022openset}
Sagar Vaze, Kai Han, Andrea Vedaldi, and Andrew Zissermann.
\newblock Open-set recognition: A good closed-set classifier is all you need?
\newblock In \emph{International Conference on Learning Representations (ICLR)}, 2022.

\bibitem[Verma et~al.(2019)Verma, Lamb, Beckham, Najafi, Mitliagkas, Lopez-Paz, and Bengio]{verma2019manifold}
Vikas Verma, Alex Lamb, Christopher Beckham, Amir Najafi, Ioannis Mitliagkas, David Lopez-Paz, and Yoshua Bengio.
\newblock Manifold mixup: Better representations by interpolating hidden states.
\newblock In \emph{International Conference on Machine Learning (ICML)}. PMLR, 2019.

\bibitem[Wang et~al.(2025)Wang, Vaze, and Han]{wang2025dissecting}
Hongjun Wang, Sagar Vaze, and Kai Han.
\newblock Dissecting out-of-distribution detection and open-set recognition: A critical analysis of methods and benchmarks.
\newblock \emph{International Journal of Computer Vision (IJCV)}, 133\penalty0 (3), 2025.

\bibitem[Wang et~al.(2022)Wang, Xu, Yang, He, Cao, and Huang]{wang2022openauc}
Zitai Wang, Qianqian Xu, Zhiyong Yang, Yuan He, Xiaochun Cao, and Qingming Huang.
\newblock Openauc: Towards auc-oriented open-set recognition.
\newblock In \emph{Advances in Neural Information Processing Systems (NeurIPS)}, 2022.

\bibitem[Wilson et~al.(2023)Wilson, Fischer, Dayoub, Miller, and S{\"u}nderhauf]{wilson2023safe}
Samuel Wilson, Tobias Fischer, Feras Dayoub, Dimitry Miller, and Niko S{\"u}nderhauf.
\newblock {SAFE}: Sensitivity-aware features for out-of-distribution object detection.
\newblock In \emph{International Conference on Computer Vision (ICCV)}, 2023.

\bibitem[Yang et~al.(2024)Yang, Zhou, Li, and Liu]{yang2024ood}
Jingkang Yang, Kaiyang Zhou, Yixuan Li, and Ziwei Liu.
\newblock Generalized out-of-distribution detection: A survey.
\newblock \emph{International Journal of Computer Vision (IJCV)}, 132\penalty0 (12), 2024.

\bibitem[Zhang et~al.(2018)Zhang, Cissé, Dauphin, and Lopez-Paz]{zhang2018mixup}
Hongyi Zhang, Moustapha Cissé, Yann~N. Dauphin, and David Lopez-Paz.
\newblock mixup: Beyond empirical risk minimization.
\newblock In \emph{International Conference on Learning Representations (ICLR)}, 2018.

\bibitem[Zhou et~al.(2021)Zhou, Ye, and Zhan]{zhou2021placeholders}
Da-Wei Zhou, Han-Jia Ye, and De-Chuan Zhan.
\newblock Learning placeholders for open-set recognition.
\newblock In \emph{Conference on Computer Vision and Pattern Recognition (CVPR)}, 2021.

\end{thebibliography}

\newpage
\appendix
\section{Supplemental}

\subsection{Method Selection and Background}
\label{sec:method-selection}

An overall view of the two-stage processing pipeline, including the nomenclature as used in this section, is given in \fig{framework}.
Below, we provide details for the representation learning and postprocessors that we investigate in this work as well as a discussion on method selection to answer our research questions.

\subsubsection{Representation Learning Methods}
\label{sec:representation-learning}

\paragraph{Methods Selection.}
Our model selection for representation learning aims to cover a representative set of methods that reflect the current state of the art in OSR as well as OOD detection while simultaneously adhering to a set of constraints given by our rigorous experimental setup to ensure that we can adequately test our research questions and hypotheses.
We selected representation learning methods that satisfy the following criteria:
\begin{enumerate}
  \item We require methods that utilize no or only natural known unknown samples as this is at the core of our analysis.
  \item Our modularization of OSR methods requires classification-based (discriminative) representation learning methods since most postprocessors rely on output logits or probabilities to perform classification.
  \item representation learning methods must be able to be trained from scratch on the ImageNet protocols without pre-training in order to prevent information leakage from training on classes that our protocols declare as unknown.
\end{enumerate}
Following, we discuss each method and its formalization in the two-stage framework in more detail.
For a systematic overview including method characteristics, see \tab{learning_methods}.

\paragraph{CE.}
Our baseline \citep{hendrycks2017baseline} training-based approach is the categorical Cross-Entropy (CE) loss trained only on samples from $K$ known classes.
For an input sample $(\sample_n, \clabel_n)\in\knowns$ and an arbitrary backbone network we obtain the deep features $\df_n\in\mathbb{R}^D$ for some deep feature dimension $D$.
These features are then passed through a fully-connected \emph{logit layer} $\logitW\in\mathbb{R}^{C\times D}$ with $C=K$ output \emph{logits} $\logits_n=\logitW\df_n\in\mathbb{R}^C$.\footnote{\label{fn:logit-cosine}Note that we can express the logit for class $c$ through the angle of the feature to the class center $\logitW_c$ as $\logits_{n,c} = \logitW\df_n = \|\df_n\| \|W_c\| \cos(\dfangle)$, where $\logitW_c$ is the $c$-th row vector of $\logitW$ and $\dfangle$ is the angle between $\df_n$ and $\logitW_c$.}
The logits are then turned into probabilities $\confidence_n\in\mathbb{R}^K$ through SoftMax activation:
\begin{equation}
  \label{eq:softmax}
  y_{n,c} = \frac{e^{z_{n,c}}}{\sum\limits_{c'=1}^C e^{z_{n,c'}}}\,.
\end{equation}
Based on these, the CE loss is computed as:
\begin{equation}
  \label{eq:ce}
  \loss_{\mathrm{CE}} = -\indicator_{(\sample_n,\clabel_n)\in\knowns}\log y_{n,\clabel_n}\,.
\end{equation}

%

\paragraph{OE.}
We include Outlier Exposure (OE) \citep{hendrycks2019oe} as state-of-the-art representation learning method from the OOD detection literature that utilize known unknown data \citep{wang2025dissecting}.
OE adds a regularization term to \eqref{eq:ce} that maximizes the entropy for known unknown samples by computing the CE loss between the uniform distribution and the SoftMax confidences of the network:
\begin{equation}
  \label{eq:oe}
  \loss_{\mathrm{OE}} = \loss_{\mathrm{CE}} - \lambda_\mathrm{OE}\, \indicator_{(\sample_n,\tau_n)\in\aux}\ \frac{1}{C}\sum_{c=1}^{C}\log y_{n,c}
\end{equation}
OE essentially is equivalent to the Entropic Open-Set (EOS) loss $\loss_{\mathrm{EOS}}$ proposed by \cite{dhamija2018agnostophobia}, with the only exception that OE provides a more intuitive way to weight the impact of known unknown samples to the training (with $\lambda_{\mathrm{OE}} = 0.5$ for computer vision tasks), whereas EOS does so   via class weighting (set to 1).
From \cite{dhamija2018agnostophobia} we know that EOS and, by extension, OE \emph{implicitly} manipulate feature magnitudes, by encouraging the network to learn small feature magnitudes for known unknown samples and large magnitudes for known samples.

\paragraph{OS.}
We employ the ObjectoSphere (OS) loss \citep{dhamija2018agnostophobia} as an extension to the EOS loss which \emph{explicitly} manipulates feature magnitudes.
It learns vanishing vectors for known unknown samples and large magnitudes for known samples by using the following regularization term combined with the EOS loss:
\begin{equation}
  \label{eq:os}
  \loss_{\mathrm{OS}} = \loss_{\mathrm{EOS}} + \lambda_\mathrm{OS} \begin{cases}
    \max(0, \xi - \|\df_n\|^2_2) & \text{if } \sample_n\in\knowns\\
    \|\df_n\|^2_2 & \text{if } \sample_n\in\aux
  \end{cases}
\end{equation}
where the hyperparameter $\xi$ is the lower bound for the feature magnitude of known samples.

\paragraph{\garbage{}.}
We use a \garbage{} classifier as representation learning baseline for OSR and OOD methods that utilize known unknown data.
\garbage{} uses the known data \knowns{} as in CE, and known unknown data \aux{} to train an additional output node $z_{n,K+1}$, so that we have a total of $C=K+1$ outputs, creating a default class for all known unknown and unknown samples.
It is trained with the standard CE loss \eqref{eq:ce} with label $\clabel=K+1$ for known unknown samples.
While this is a common approach in object detection models \citep{dhamija2020elephant}, which collect a lot of background samples, it is only rarely applied directly to OSR problems \citep{dhamija2018agnostophobia,palechor2023protocols} or as component of more complex architectures such as PROSER or G-OpenMax \citep{zhou2021placeholders,ge2017generative}. 
Depending on the context, a \garbage{} classifier is known as \emph{Background Class} \citep{dhamija2020elephant,dhamija2018agnostophobia,palechor2023protocols} or \emph{Dummy Classifier} \citep{zhou2021placeholders}.
Note that for computing OOD scores $\score_n$ (see \ref{sec:postprocessing}) for the \garbage{} classifier, we follow \citet{ren2019likelihood} and ignore the unknown class ($y_{n,K+1}$ or $z_{n,K+1}$) if it exists.
A low probability for the unknown class $y_{n,K+1}$ does not indicate a high probability for any of the known classes.
On the other hand, due to Softmax requiring probabilities sum up to 1, a large probability $y_{n,K+1}$ enforces low probabilities for all known classes.
Therefore, $y_{n,K+1}$ does not add any new information.

\paragraph{ARPL.}
Finally, we include the Adversarial Reciprocal Point Learning (ARPL) loss \citep{chen2021arpl}, which currently is the state of the art for OSR\footnote{Note that we do not use ARPL+CS with the generator for confusing samples (CS), since it is prohibitively expensive to train at large scale \citep{vaze2022openset}.} and trained solely on the known data \knowns.
Unlike most training-based methods, ARPL does not learn a prototype for a class $c$, but a reciprocal point $\mathbf p_c\in\mathbb{R}^D$ in deep feature space, \ie, a point that represents everything \emph{but} class $c$.
The ARPL objective maximizes the distance between the reciprocal point and the features of samples from the respective class by computing logits $\logits_n$ via distance measures between deep features $\df_n$ and the reciprocal points $\mathbf p_c$:
\begin{equation}
  z_{n,c} = \frac{1}{D}\|\df_n - \mathbf p_c\|^2_2 - \df_n\T \mathbf p_c\,,
\end{equation}
which are used via SoftMax \eqref{eq:softmax} in the CE loss \eqref{eq:ce}.
To constrain open space, the distance between deep features and reciprocal points is bound by a learnable constant $\rho$ via the regularization term with weight $\lambda_{\mathrm{ARPL}}=0.1$:
\begin{equation}
  \label{eq:arpl-reg}
  \loss_{\mathrm{ARPL}} = \loss_{\mathrm{CE}} + \lambda_{\mathrm{ARPL}} \indicator_{(\sample_n,\tau_n)\in\knowns} \max\left(0, \frac{1}{D}\|\df_n - \mathbf p_{\tau_n}\|^2_2 - \rho\right)
\end{equation}

\subsubsection{Postprocessing Methods}
\label{sec:postprocessing}

\paragraph{Method Selection.}
We selected postprocessors to cover a split between representative magnitude-aware and magnitude-unaware methods.
Furthermore, we aim to cover both trainable and non-trainable methods to include methods that can specifically adjust to differently learned representations.
Following, we discuss each method and its formalization in the two-stage framework in more detail.
For a systematic overview including method characteristics, see \tab{postprocessing_methods}.

\paragraph{MSP.}
The Maximum Softmax Probability (MSP) \citep{hendrycks2017baseline} is the de-facto default postprocessor for OSR and OOD detection and serves as our baseline.
Class predictions and OOD scores are computed from probabilities for known classes as:
\begin{equation}
  \label{eq:msp}
  \pred_n = \argmax_{1\leq c\leq K} y_{n,c} \qquad \text{and} \qquad \score_n = y_{n,\pred_n}\,.
\end{equation}
MSP is not magnitude-aware since SoftMax \eqref{eq:softmax} normalizes the logits and only considers their relative differences.

\paragraph{MaxLogits.}
MaxLogits \citep{hendrycks2022scaling} or MLS \citep{vaze2022openset} exploit the magnitude of the logits $\logits_n$, which contains useful information for OSR and OOD detection that is lost during softmax.
MLS is magnitude-aware since logit magnitude is linked to feature magnitude \citep{wang2025dissecting}.\footref{fn:logit-cosine}
Class predictions and OOD scores are computed from known logits as:
\begin{equation}
  \label{eq:maxlogits}
  \pred_n = \argmax_{1\leq c\leq K} z_{n,c} \qquad \text{and} \qquad \score_n = z_{n,\pred_n}\,.
\end{equation}

\paragraph{OpenMax} \citep{bendale2016openmax} uses deep features $\df_n$, referred to as Activation Vectors (AVs), to statistically model probabilities for an additional output node for the unknown class.
During training, for each known class $c$, the Mean Activation Vector (MAV) $\mav_c$ is computed by averaging the deep features $\df_n$ extracted from all correctly classified training samples.
Then, the cosine distances of the MAV $\mav_c$ to all the AVs $\df_{n,c}$ of the same class are computed.
Here, we make use of a twist implemented in the VAST package of the original authors:\footnote{\label{fn:vast}\url{https://github.com/Vastlab/vast}} instead of using the original distances to model the distribution, we multiply the cosine distances by a factor $\kappa$, which allows modeling more compact class representations:
\begin{equation}
  \label{eq:openmax-distances}
  d_{n,c} = \kappa(1-\cos(\df_n, \mav_c))
\end{equation}
Then, a per-class Weibull distribution $\Psi_c$ is fitted to the top $\lambda$ largest cosine distances $d_{n,c}$.
For features $\df_n$ of a test sample, the class-wise Weibull distributions estimate a logit $\hat{z}_{n,K+1}$ for the unknown class, as well as modifying the logits for the top $\alpha$ classes, giving newly estimated logits $\hat{z}_{n,c}$.
From these, the output $\hat y_{n,c}$ is computed via softmax \eqref{eq:softmax} and the output $\scores_n$ is computed as:
\begin{equation}
  \pred_n = \argmax_{1\leq c\leq K} \hat{z}_{n,c} \qquad \text{and} \qquad \score_n = \hat{y}_{n,\pred_n}\,.
\end{equation}
OpenMax is not magnitude-aware since the cosine distance ignores the feature magnitude.

\paragraph{PostMax.}
Postnormalization of Maxima (PostMax) \citep{cruz2024postmax} uses Extreme Value Theory (EVT) by applying a Generalized Pareto Distribution (GPD) to maximum logits post-normalized by the feature magnitude.
Based on their findings that unknown samples have larger feature magnitude than known samples on large-scale data, they normalize logits by dividing them by their deep feature magnitude to further increase the separation between known and unknown samples:
\begin{equation}
  \label{eq:postmax-norm}
  \hat{\logits}_n = \frac{\logits_n}{\|\df_n\|_2+1}\,.
\end{equation}
This makes it an explicitly magnitude-aware method.
We modify the original implementation to shift the magnitudes by 1 to avoid issues with magnitudes $\|\df\|_2<1$ which reverse the desired effect of normalization.\footnote{We also tried different normalization techniques, including to multiply with the norm, which seems more reasonable for MM representation learning methods. However, the detrimental effects of MM representation learning with PostMax for large-scale evaluations was present in any case. This modification does not harm performance with non-MM representation learning methods.}
This occurs consistently for features from magnitude-manipulating representation learning methods, but not for others.
The class-agnostic GPD $\Psi_{\mu,\sigma,\xi}$ is fitted only on the maximum normalized logits of correctly classified known training samples, which is then used to compute a probability of the sample being known.
Following \cite{cruz2024postmax}, the scores $\scores_n$ are computed as:
\begin{equation}
  \pred_n = \argmax_{1\leq c\leq K} \hat{z}_{n,c} \qquad \text{and} \qquad \score_n = \Psi_{\mu,\sigma,\xi}(\hat{z}_{n,\pred_n})\,.
\end{equation}

\paragraph{GHOST.}
The Gaussian Hypothesis Open-Set Technique (GHOST) \citep{rabinowitz2025ghost} models each class in deep feature space as a multivariate Gaussian distribution with the intuition that features $\df$ from known and unknown samples deviate by feature magnitude even if the angular direction overlaps, making it magnitude-aware.
During training, GHOST fits a Gaussian distribution $(\mu_c,\sigma_c)$ for each known class $c$ based on the deep features $\df_n$ of correctly classified training samples.
During inference, GHOST first computes a class prediction from the logits:
\begin{equation}
  \pred_n = \argmax_{1\leq c\leq K} z_{n,c}\,,
\end{equation}
as well as z-score $s_n$ for each sample $\sample_n$ and corresponding Gaussian $(\mu_{\pred_n},\sigma_{\pred_n})$.
This is then used to compute the score $\score_n$ by dividing the original logit:
\begin{equation}
  \label{eq:ghost-zscore}
  s_n = \sum_{d=1}^{D} \frac{|\df_{n,d} - \mu_{\pred_n,d}|}{\sigma_{\pred_n,d}} , \qquad \score_n = \frac{z_{n,\pred_n}}{s_n}\,.
\end{equation}

\subsection{Derivation of Performance Decomposition}

Here we provide the derivation for the performance decomposition of the performance increase of a \comb{i}{j} over \comb{CE}{MSP} into the sum of the individual performance gains of the representation learning method and postprocessor.
For the decomposition as in Equation \eqref{eq:perfrmance-decomposition-msp} we have:
We start with the performance of \comb{i}{j} and add and subtract the baseline \comb{CE}{MSP} and the intermediate \comb{i}{MSP}:

\begin{align}
  \metric(\comb{i}{j}) &= \metric(\comb{i}{j}) + \underbrace{ \metric(\comb{CE}{MSP}) - \metric(\comb{CE}{MSP}) }_{=0} \\
  &\quad + \underbrace{ \metric(\comb{i}{MSP}) - \metric(\comb{i}{MSP}) }_{=0} \\
  \intertext{By rearanging and using definitions of the contribution deltas \deltarl{} and \deltapp{} (Equations \eqref{eq:performance-delta-rl} and \eqref{eq:performance-delta-pp}) we have:}
  &= \metric(\comb{CE}{MSP}) + \underbrace{ (\metric(\comb{i}{MSP}) - \metric(\comb{CE}{MSP})) }_{= \deltarl[MSP]{i}} \\
  &\quad + \underbrace{ (\metric(\comb{i}{j}) - \metric(\comb{i}{MSP})) }_{= \deltapp[i]{j}} \\
  \intertext{By rearranging we have:}
  &\iff \metric(\comb{i}{j}) - \metric(\comb{CE}{MSP}) = \deltarl[MSP]{i} + \deltapp[i]{j}
\end{align}

Analogously for the decomposition as in Equation \eqref{eq:perfrmance-decomposition-ce} we have:
We start with the performance of \comb{i}{j} and add and subtract the baseline \comb{CE}{MSP} and the intermediate \comb{CE}{j}:

\begin{align}
  \metric(\comb{i}{j}) &= \metric(\comb{i}{j}) + \underbrace{ \metric(\comb{CE}{MSP}) - \metric(\comb{CE}{MSP}) }_{=0} \\
  &\quad + \underbrace{ \metric(\comb{CE}{j}) - \metric(\comb{CE}{j}) }_{=0} \\
  \intertext{By rearanging and using definitions of the contribution deltas \deltarl{} and \deltapp{} (Equations \eqref{eq:performance-delta-rl} and \eqref{eq:performance-delta-pp}) we have:}
  &= \metric(\comb{CE}{MSP}) + \underbrace{ (\metric(\comb{i}{j}) - \metric(\comb{CE}{j})) }_{= \deltarl[j]{i}} \\
  &\quad + \underbrace{ (\metric(\comb{CE}{j}) - \metric(\comb{CE}{MSP})) }_{= \deltapp[CE]{j}} \\
  &\iff \metric(\comb{i}{j}) - \metric(\comb{CE}{MSP}) = \deltarl[j]{i} + \deltapp[CE]{j}
\end{align}

\subsection{Derivation for Feature Magnitude Incentive}
\label{sec:derivation_feature_magnitude}

Here, we provide the derivation for the training incentive of CE with $C=K$ (or $C = K+1$) to learn ``sufficiently large'' feature magnitudes:
\begin{equation*}
  {\|\df\|}_2 > \frac{\lb}{\delta_{\logits_c/ {\|\df\|}_2}} = \frac{\log(\frac{1-\epsilon}{\epsilon}) + \log(C-1)}{{\|\logitW_c\|}_2 \cos(\dfangle_{c}) - \max_{c^\prime\neq c}\{{\|\logitW_{c^\prime}\|}_2 \cos(\dfangle_{c^\prime})\}} \Longrightarrow \confidence_c > 1-\epsilon \,,
\end{equation*}

For some small $\epsilon>0$ and class $c\in\{1,\ldots,C\}$, we want the trained network to be able to achieve the maximum probability over all classes $\confidence_c > 1-\epsilon$ for any sample.
Note that we have $\confidence_c = \max_{c^\prime\in\{1,\ldots, C\}}\confidence_{c^\prime}$.
\begin{align*}
  \confidence_c > 1-\epsilon &\iff \frac{e^{\logits_c}}{\sum_{c^\prime} e^{\logits_{c^\prime}}} = \frac{e^{\logits_c}}{e^{\logits_c} + \sum_{c^\prime\neq c} e^{\logits_{c^\prime}}} > 1-\epsilon \\
  &\iff e^{\logits_c} > (1-\epsilon) (e^{\logits_c} + \sum_{c^\prime\neq c} e^{\logits_{c^\prime}}) \\
  &\iff e^{\logits_c} > e^{\logits_c} + \sum_{c^\prime\neq c} e^{\logits_{c^\prime}} + \epsilon e^{\logits_c} - \epsilon \sum_{c^\prime\neq c} e^{\logits_{c^\prime}} \\
  &\iff e^{\logits_c} > \frac{1-\epsilon}{\epsilon} \sum_{c^\prime\neq c} e^{\logits_{c^\prime}} \\
  &\iff \logits_c > \underbrace{\log\left(\frac{1-\epsilon}{\epsilon}\right)}_{=:\beta} + \log\left(\sum_{c^\prime\neq c} e^{\logits_{c^\prime}}\right)
\end{align*}
By binding the LogSumExp $\log\left(\sum_{c^\prime\neq c} e^{\logits_{c^\prime}}\right)$ above, we get a sufficient condition $\delta_{\logits_c} > \lb$ such that $\delta_{\logits_c} > \lb \Longrightarrow \confidence_c > 1-\epsilon$.
We have the upper bound of the LogSumExp function:
\begin{equation*}
  \log\left(\sum_{c^\prime\neq c} e^{\logits_{c^\prime}}\right) \leq \log(C-1) + \max_{c\neq c^\prime} \logits_{c^\prime}
\end{equation*}
\begin{equation*}
  \logits_c > \beta + \max_{c\neq c^\prime} \logits_{c^\prime} + \log(C-1) \Longrightarrow \logits_c > \beta + \log\left(\sum_{c^\prime\neq c} e^{\logits_{c^\prime}}\right)
\end{equation*}
From this we get:
\begin{align*}
  \logits_c > \beta + \max_{c\neq c^\prime} \logits_{c^\prime} + \log(C-1) &\iff \underbrace{\logits_c - \max_{c\neq c^\prime} \logits_{c^\prime}}_{=: \delta_{\logits_c}} > \beta + \log(C-1) =: \lb
\end{align*}
This provides a lower bound on the difference between the maximal two logits. Using footnote \ref{fn:logit-cosine} we have
\begin{align*}
  &\iff {\|\df\|}_2{\|\logitW_c\|}_2 \cos(\dfangle_{\df,\logitW_c}) - \max_{c^\prime \neq c} \{{\|\df\|}_2{\|\logitW_{c^\prime}\|}_2 \cos(\dfangle_{\df,\logitW_{c^\prime}})\} > \lb \\
  &\iff {\|\df\|}_2\underbrace{({\|\logitW_c\|}_2 \cos(\dfangle_{\df,\logitW_c}) - \max_{c^\prime \neq c} {\|\logitW_{c^\prime}\|}_2 \cos(\dfangle_{\df,\logitW_{c^\prime}})\})}_{=:\delta_{{\logits_c}/{\|\df\|}_2}} > \lb \\
  &\iff {\|\df\|}_2 > \frac{\lb}{\delta_{{\logits_c}/{\|\df\|}_2}}
\end{align*}
where ${\delta_{{\logits_c}/{\|\df\|}_2}} > 0$.

\subsection{EMNIST Benchmark}
\label{sec:emnist}
We perform qualitative small-scale experiments on the \textbf{EMNIST} protocol \citep{dhamija2018agnostophobia} to replicate settings of high similarity between known and known unknown classes at small scale.
This benchmark contains a wide spectrum of clearly attributable visual similarities between known and known unknown classes because the samples do not contain any image background.
It consists of EMNIST MNIST split as knowns, the first half of EMNIST letters (a-m) as known unknown data, and the second half (n-z) as unknowns.\footnote{Contrary to \citep{dhamija2018agnostophobia} we use EMNIST MNIST instead of the original MNIST because the preprocessing, while similar, is not exact \citep{cohen2017emnist}. EMNIST MNIST and letters contain noticeably softer and thicker digits than MNIST. In order to focus on semantic shift and not introduce easily learnable covariate shift through softness and sharpness, we use EMNIST MNIST.}
While most digits (known) do not contain any visually similar letters, digits 1 and 9 contain known unknown classes (``i'', ``l'' and ``g'', ``q'' respectively) that are visually very similar or indistinguishable without context depending on handwriting and capitalization.

To visualize how magnitude collapse and known class classification interacts, we repeat the EMNIST experiments with a 2D bottleneck deep feature layer, following \citet{dhamija2018agnostophobia}.
These can illustrate how different representation learning methods structure the feature space and how this affects postprocessors, however, the low dimensionality limits not only the representational capacity but also changes the geometry of the feature space and the behavior, so extrapolation to high dimensional feature spaces is limited.
Experiments are otherwise identical to those in \sec{experimental-setup}.

\begin{figure}
  \centering
  \includegraphics[width=\textwidth-0.05\textwidth]{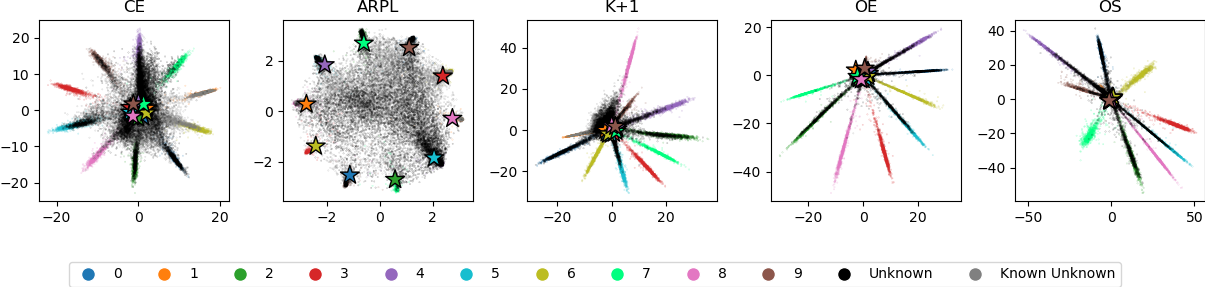}
  \Caption{Feature Visualization}{We provide 2D feature visualizations of EMNIST experiments with different representation learning methods.}
\end{figure}

\subsection{Hyperparameter Optimization}
\label{sec:hyperparameter-optimization}

For most methods, we rely on the hyperparameters as provided by the according papers.
Few methods however do not provide any, namely: $\lambda$ and $\xi$ for ObjectoSphere and $\lambda$, $\kappa$, and $\alpha$ for OpenMax, where we perform hyperparameter optimization on the validation set using grid search, based on the maximum AUOSCR.
For OpenMax the parameter optimization is performed for each upstream representation learning method separately to ensure optimal settings, since the feature representations differ significantly.
The considered hyperparameter ranges are as follows:
\begin{itemize}[leftmargin=1.5em]
  \item OS: $\lambda\in\{10^{-4},10^{-3},10^{-2}\}$, $\xi\in\{10,20,30\}$
  \item OpenMax: $\alpha\in\{1,2,3,5,10\}$, $\kappa\in\{1.5,1.7,2,2.3\}$, $\lambda\in\{10,50,100,250,500,750,1000\}$
\end{itemize}
The best hyperparameter values for each method and protocol are summarized in \tab{hyperparameters}.

\begin{table*}[t]
\centering
\scalebox{0.70}{
  \begin{tabular}{@{}lcccccc@{}}
  \toprule
  & & \multicolumn{2}{c}{CIFAR+N} & \multicolumn{3}{c}{ImageNet} \\
  Method & Parameter & CIFAR+10 & CIFAR+50 & $P_1$ & $P_2$ & $P_3$ \\
  \midrule
  \rl{OS} & $\lambda_{\mathrm{OS}}$ & $10^{-3}$/$10^{-2}$/$10^{-2}$/$10^{-2}$/$10^{-2}$ & $10^{-3}$/$10^{-2}$/$10^{-2}$/$10^{-3}$/$10^{-2}$ & $10^{-3}$ & $10^{-3}$ & $10^{-3}$ \\
  & $\xi$ & 10/10/10/10/10 & 20/10/10/10/10 & 30 & 20 & 10 \\
  \midrule
  \comb{CE}{OpenMax} & $\alpha$ & 3/3/3/1/3 & 3/2/3/3/3 & 5 & 5 & 2 \\
  & $\kappa$ & 1.5/1.5/2.3/2/1.7 & 1.7/1.5/2.3/1.5/2.3 & 2.3 & 2.3 & 2 \\
  & $\lambda$ & 100/100/250/10/100 & 250/100/500/100/100 & 750 & 750 & 100 \\
  \midrule
  \comb{ARPL}{OpenMax} & $\alpha$ & 3/3/1/3/3 & 3/2/3/3/3 & 5 & 3 & 2 \\
  & $\kappa$ & 1.5/2.3/2/1.5/1.7 & 2/1.5/1.5/1.5/1.5 & 2.3 & 2.3 & 2.3 \\
  & $\lambda$ & 100/250/10/100/100 & 100/100/100/100/100 & 750 & 750 & 10 \\
  \midrule
  \comb{\garbage{}}{OpenMax} & $\alpha$ & 1/1/1/1/1 & 1/1/1/1/1 & 5 & 3 & 5 \\
  & $\kappa$ & 1.5/2/2.3/1.7/2 & 1.7/2/1.5/1.5/2.3 & 1.7 & 2.3 & 2 \\
  & $\lambda$ & 100/250/100/100/100 & 250/500/250/250/250 & 750 & 750 & 100 \\
  \midrule
  \comb{OE}{OpenMax} & $\alpha$ & 1/2/1/3/3 & 3/3/1/2/1 & 3 & 2 & 2 \\
  & $\kappa$ & 1.5/1.7/2/2/2 & 2.3/1.5/1.5/2/1.5 & 1.7 & 1.5 & 2.3 \\
  & $\lambda$ & 10/100/10/100/250 & 100/100/10/100/10 & 10 & 10 & 10 \\
  \midrule
  \comb{OS}{OpenMax} & $\alpha$ & 1/1/3/2/2 & 3/2/3/3/2 & 10 & 5 & 10 \\
  & $\kappa$ & 1.5/2.3/1.5/1.5/2.3 & 2.3/1.7/2/2.3/2 & 2.3 & 2.3 & 2 \\
  & $\lambda$ & 10/10/10/10/10 & 100/10/10/100/10 & 10 & 10 & 10 \\
  \bottomrule
  \end{tabular}
}
\Caption[tab:hyperparameters]{Hyperparameters}{This table include the optimized hyperparameter values for each method and protocol. CIFAR+N results are reported for each trial separately.}
\end{table*}

\subsection{Performance Metrics}
\label{sec:performance-metrics}
\paragraph{CCR@FPR}
computes the CCR at a specific FPRs $\fprval \in \{10^{-4}, 10^{-3}, 10^{-2}, 10^{-1}, 10^0\}$ and provides insights into the classification performance at various tolerances for errors caused by missed rejections.
It is therefore highly relevant for practical applications where a fixed threshold $\threshold$ is required, which is typically selected based on a certain FPR.
It is computed as:

\begin{align}
  \label{eq:ccr_at_fpr}
  \mathrm{CCR@FPR} = \begin{cases} \mathrm{CCR}(\threshold_\fprval) & \text{if } \threshold_\fprval \text{ exists} \\ 0 & \text{otherwise} \end{cases}\,,
\end{align}
where $\threshold_\fprval = \mathrm{FPR}^{-1}(\fprval)$ is the threshold that yields $\mathrm{FPR}=\fprval$.
CCR@FPR for $\fprval=10^0$ resembles closed-set accuracy.

\paragraph{AUOSCR.}
The OSCR curve \citep{dhamija2018agnostophobia} simultaneously evaluates classification of known samples via CCR as well as the rejection of unknown samples via FPR over all possible thresholds.
It is computed by varying the threshold $\threshold$ from the smallest to the largest possible score value, and plotting the CCR over the FPR, \ie, computing CCR@FPR at all thresholds.
The Area Under the OSCR curve (AUOSCR) is computed by integrating the OSCR curve from $\fprval=0$ to $\fprval=1$.
Since the OSCR curve is a monotonically increasing function, the AUOSCR is maximized at and bounded by the closed-set accuracy.

\paragraph{OOSA.}
The Operational Open-set Accuracy (OOSA) \citep{cruz2024postmax} evaluates the open-set performance at a fixed operational threshold $\threshold^\star$, determined on the validation set, and provides insights into the performance of a method in a real-world setting.
It is defined as a trade-off between the CCR and the Unknown Rejection Rate (URR), $\mathrm{URR}(\threshold)=1-\mathrm{FPR}(\threshold)$:
\begin{equation}
  \label{eq:oosa}
  \mathrm{OOSA} = \alpha_{\mathrm{CCR}}\mathrm{CCR}(\threshold^\star) + (1-\alpha_{\mathrm{CCR}})\mathrm{URR}(\threshold^\star)
\end{equation}
where $\threshold^\star$ is the operational threshold that maximizes this equation on the validation set.
We follow \citet{cruz2024postmax} and set $\alpha_{\mathrm{CCR}}=\frac{\mid \knowns_\textrm{test} \mid}{\mid \knowns_\textrm{test} \mid + \mid \unknowns_\textrm{test} \mid}$ to equally weight known and unknown test samples.

\paragraph{AUROC.}
In order to evaluate the OOD detection capabilities independently of the ID classification task, we use the Area Under the Receiver Operating Characteristics (AUROC) curve \citep{hendrycks2017baseline,hendrycks2019oe,chen2021arpl,hendrycks2022scaling,vaze2022openset,yang2024ood,wang2025dissecting}.
AUROC concerns how well known and unknown classes can be distinguished by computing FPR \eqref{eq:oscr}, as well as the True Positive Rate (TPR):

\begin{equation}
  \label{eq:tpr}
  \mathrm{TPR}(\threshold) = \frac{\bigl|\{(\sample_n,\clabel_n)\in\knowns \wedge \score_n \geq \threshold \}\bigr|}{N_K}
\end{equation}

The ROC is computed by varying $\threshold$, and the area under that curve is determined.

\begin{figure}[ht]
  \centering
  \includegraphics[width=.7\textwidth-0.05\textwidth]{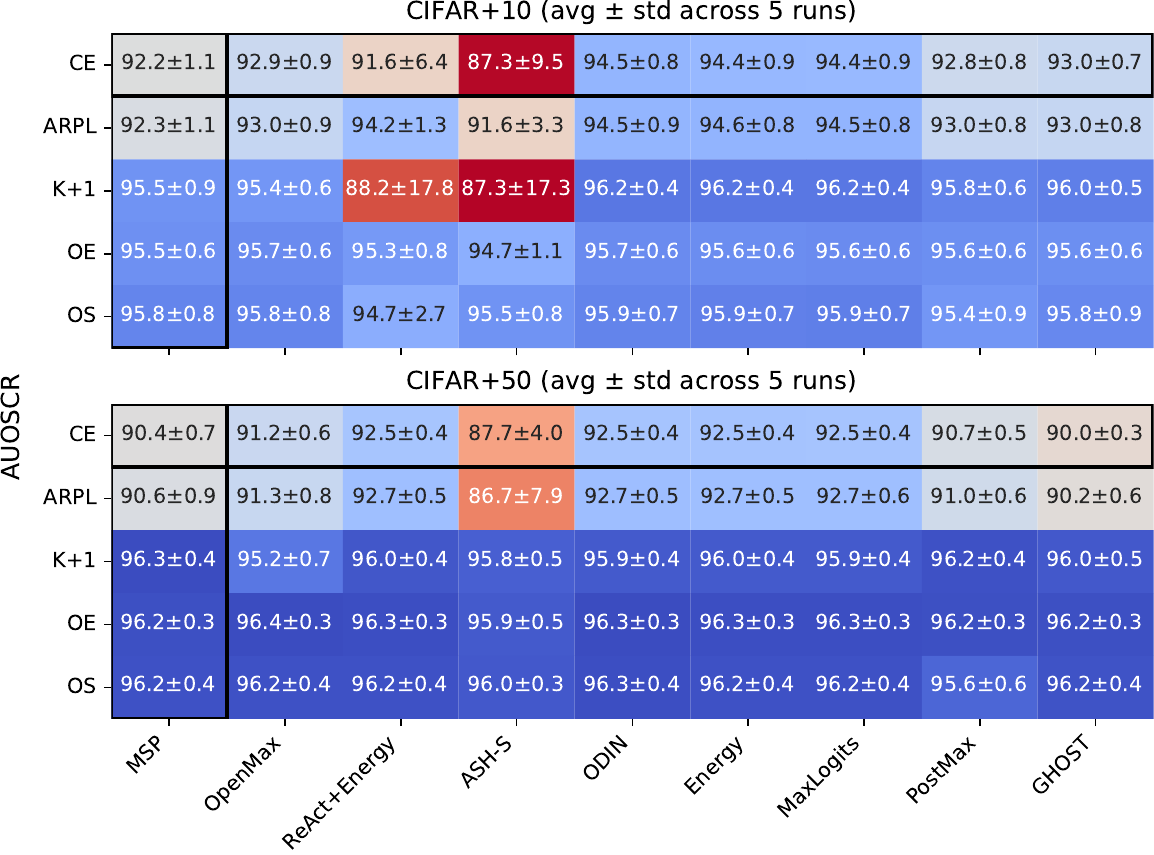}
  \Caption[fig:supplemental-score-heatmaps-cifar]{CIFAR+N OSR Performance}{
    We show OSR performance for \comb{\rl}{\pp} combinations across datasets in AUOSCR.
    The color of each heatmap is normalized independently and centered at the \comb{CE}{MSP} baseline, where blue shows an increase and red a decrease. Single-stage methods are surrounded by a black border.
    Results depict the average performance $\pm$ standard deviation over 5 trials.
  }
\end{figure}

\begin{figure}[ht]
  \centering
  \includegraphics[width=\textwidth-0.05\textwidth]{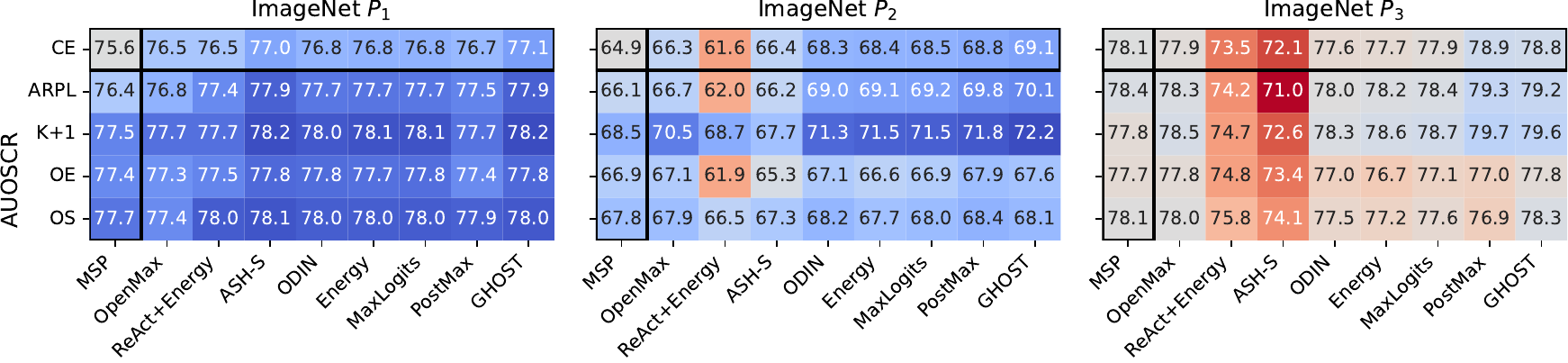}
  \Caption[fig:supplemental-score-heatmaps-imagenet]{ImageNet OSR Performance}{
    We show OSR performance for \comb{\rl}{\pp} combinations across datasets in AUOSCR.
    The color of each heatmap is normalized independently and centered at the \comb{CE}{MSP} baseline, where blue shows an increase and red a decrease. Single-stage methods are surrounded by a black border.
  }
\end{figure}

\subsection{Experiments on Additional Postprocessing Methods}

To assess the generalizability of our findings, we conducted additional experiments on four widely used postprocessing methods that were not included in the main paper.
Particularly, we include Energy \citep{liu2020energy}, ODIN \citep{liang2017odin}, ReAct+Energy \citep{sun2021react}, and ASH-S \citep{djurisic2023-ash}.
The AUOSCR heatmaps for combinations on CIFAR+N and ImageNet are shown in \fig{supplemental-score-heatmaps} and \fig{supplemental-score-heatmaps-imagenet}, respectively.
For Energy and ODIN we observe nearly identical effects as previous postprocessors, while ReAct+Energy and ASH-S struggled with high-similarity setting $P_3$ (shifting the color-map and slightly hiding the effect).
On the smaller CIFAR+N dataset, we observe that ReAct+Energy and ASH-S can exhibit high variation in performance, dragging down the average performance significantly, while other poseprocessors are highly stable.

\subsection{Additional Figures and Tables}
The main paper contained only a subset of evaluation metrics and visualizatrions.
Here we provide remaining figures and tables containing the exact results.

\begin{table*}[p]
  \centering
  \scalebox{0.60}{\begin{tabular}{@{}l@{}c*{8}{c@\ }c@{}}
\toprule
Dataset & $f$ & $g$ & CCR@0.01 & CCR@0.1 & CCR@1 & CCR@10 & CCR@100 & AUOSCR & AUROC & OOSA \\
\midrule
CIFAR+10 & CE & MSP & 43.0 $\pm$ 13.2 & 40.5 $\pm$ 13.3 & 63.8 $\pm$ 5.1 & 84.5 $\pm$ 2.4 & \textit{97.0 $\pm$ 0.7} & 92.2 $\pm$ 1.1 & 93.7 $\pm$ 0.8 & 87.8 $\pm$ 1.1 \\
 &  & OpenMax & 43.8 $\pm$ 12.6 & 44.1 $\pm$ 12.3 & 67.8 $\pm$ 3.0 & 86.1 $\pm$ 1.9 & \textit{97.0 $\pm$ 0.7} & 92.9 $\pm$ 0.9 & 94.5 $\pm$ 0.5 & 88.4 $\pm$ 1.0 \\
 &  & MaxLogits & \textit{46.5 $\pm$ 9.2} & \textit{46.5 $\pm$ 9.2} & \textit{72.5 $\pm$ 2.7} & \textit{89.8 $\pm$ 1.7} & \textit{97.0 $\pm$ 0.7} & \textit{94.4 $\pm$ 0.9} & \textit{96.6 $\pm$ 0.5} & \textit{90.5 $\pm$ 1.0} \\
 &  & PostMax & 33.6 $\pm$ 15.3 & 33.6 $\pm$ 15.3 & 64.6 $\pm$ 4.3 & 85.3 $\pm$ 1.6 & \textit{97.0 $\pm$ 0.7} & 92.8 $\pm$ 0.7 & 94.7 $\pm$ 0.3 & 88.3 $\pm$ 1.0 \\
 &  & GHOST & 3.2 $\pm$ 3.7 & 2.6 $\pm$ 3.5 & 28.1 $\pm$ 10.2 & 87.7 $\pm$ 1.3 & \textit{97.0 $\pm$ 0.7} & 93.0 $\pm$ 0.7 & 95.1 $\pm$ 0.3 & 90.0 $\pm$ 1.1 \\
\cmidrule{2-11}
 & ARPL & MSP & 41.0 $\pm$ 11.8 & 42.8 $\pm$ 9.6 & 65.3 $\pm$ 1.7 & 84.6 $\pm$ 2.1 & \textit{97.1 $\pm$ 0.7} & 92.4 $\pm$ 1.1 & 93.9 $\pm$ 0.8 & 87.9 $\pm$ 1.1 \\
 &  & OpenMax & 44.4 $\pm$ 9.4 & 44.6 $\pm$ 9.2 & 68.2 $\pm$ 1.5 & 86.4 $\pm$ 1.6 & \textit{97.1 $\pm$ 0.7} & 93.0 $\pm$ 0.9 & 94.7 $\pm$ 0.5 & 88.6 $\pm$ 0.9 \\
 &  & MaxLogits & \textit{48.8 $\pm$ 12.4} & \textit{48.8 $\pm$ 12.4} & \textit{73.2 $\pm$ 2.9} & \textit{90.2 $\pm$ 1.2} & \textit{97.1 $\pm$ 0.7} & \textit{94.5 $\pm$ 0.8} & \textit{96.8 $\pm$ 0.4} & \textit{90.7 $\pm$ 1.0} \\
 &  & PostMax & 36.3 $\pm$ 15.0 & 36.3 $\pm$ 15.0 & 64.3 $\pm$ 2.4 & 85.5 $\pm$ 1.4 & \textit{97.1 $\pm$ 0.7} & 93.0 $\pm$ 0.8 & 94.9 $\pm$ 0.4 & 88.7 $\pm$ 0.7 \\
 &  & GHOST & 3.6 $\pm$ 4.3 & 3.6 $\pm$ 4.3 & 23.3 $\pm$ 7.1 & 88.1 $\pm$ 1.4 & \textit{97.1 $\pm$ 0.7} & 93.0 $\pm$ 0.8 & 95.2 $\pm$ 0.4 & 90.2 $\pm$ 1.0 \\
\cmidrule{2-11}
 & K+1 & MSP & 58.0 $\pm$ 18.2 & 58.1 $\pm$ 18.2 & 80.7 $\pm$ 7.6 & 92.6 $\pm$ 2.5 & 97.3 $\pm$ 0.7 & 95.5 $\pm$ 0.9 & 97.9 $\pm$ 1.4 & 92.2 $\pm$ 2.5 \\
 &  & OpenMax & 61.8 $\pm$ 12.9 & 61.8 $\pm$ 12.9 & 80.8 $\pm$ 4.0 & 92.0 $\pm$ 1.6 & \textbf{\textit{97.4 $\pm$ 0.7}} & 95.4 $\pm$ 0.7 & 97.3 $\pm$ 0.9 & 91.9 $\pm$ 1.2 \\
 &  & MaxLogits & \textit{63.1 $\pm$ 22.8} & \textit{63.1 $\pm$ 22.8} & \textbf{\textit{84.2 $\pm$ 4.4}} & \textbf{\textit{94.2 $\pm$ 0.9}} & 97.3 $\pm$ 0.7 & \textbf{\textit{96.2 $\pm$ 0.4}} & \textbf{\textit{98.6 $\pm$ 0.6}} & \textbf{\textit{93.7 $\pm$ 0.7}} \\
 &  & PostMax & 57.9 $\pm$ 19.8 & 57.9 $\pm$ 19.8 & 81.3 $\pm$ 7.8 & 93.5 $\pm$ 1.9 & 97.3 $\pm$ 0.7 & 95.8 $\pm$ 0.6 & 98.3 $\pm$ 1.0 & 93.0 $\pm$ 1.8 \\
 &  & GHOST & 34.8 $\pm$ 31.4 & 34.8 $\pm$ 31.4 & 81.1 $\pm$ 7.7 & 94.1 $\pm$ 1.3 & 97.3 $\pm$ 0.7 & 96.0 $\pm$ 0.5 & 98.4 $\pm$ 0.8 & 93.6 $\pm$ 1.1 \\
\cmidrule{2-11}
 & OE & MSP & 57.4 $\pm$ 17.4 & 57.4 $\pm$ 17.3 & 81.4 $\pm$ 6.1 & 92.9 $\pm$ 1.8 & 97.1 $\pm$ 0.8 & 95.6 $\pm$ 0.6 & 98.0 $\pm$ 1.0 & 92.6 $\pm$ 1.6 \\
 &  & OpenMax & 58.2 $\pm$ 18.2 & 58.2 $\pm$ 18.2 & 81.5 $\pm$ 5.7 & 93.2 $\pm$ 1.8 & \textit{97.2 $\pm$ 0.7} & \textit{95.7 $\pm$ 0.6} & 98.0 $\pm$ 0.9 & 92.7 $\pm$ 1.5 \\
 &  & MaxLogits & \textit{58.8 $\pm$ 19.5} & \textit{58.8 $\pm$ 19.5} & \textit{82.4 $\pm$ 5.8} & \textit{93.3 $\pm$ 1.5} & 97.1 $\pm$ 0.8 & 95.6 $\pm$ 0.6 & \textit{98.2 $\pm$ 0.9} & 92.6 $\pm$ 1.6 \\
 &  & PostMax & 52.3 $\pm$ 19.6 & 52.3 $\pm$ 19.6 & 81.3 $\pm$ 6.2 & 93.2 $\pm$ 1.5 & 97.1 $\pm$ 0.8 & 95.6 $\pm$ 0.6 & 98.1 $\pm$ 0.9 & \textit{92.8 $\pm$ 1.5} \\
 &  & GHOST & 36.0 $\pm$ 32.5 & 36.0 $\pm$ 32.5 & 81.7 $\pm$ 6.4 & \textit{93.3 $\pm$ 1.5} & 97.1 $\pm$ 0.8 & 95.6 $\pm$ 0.6 & 98.1 $\pm$ 1.0 & 92.7 $\pm$ 1.6 \\
\cmidrule{2-11}
 & OS & MSP & \textbf{\textit{65.5 $\pm$ 14.8}} & \textbf{\textit{65.5 $\pm$ 14.8}} & 82.1 $\pm$ 6.5 & 93.2 $\pm$ 2.0 & \textbf{\textit{97.4 $\pm$ 0.7}} & 95.8 $\pm$ 0.8 & 98.0 $\pm$ 1.1 & \textit{92.7 $\pm$ 1.9} \\
 &  & OpenMax & 65.0 $\pm$ 17.7 & 65.0 $\pm$ 17.7 & 81.9 $\pm$ 6.9 & 93.3 $\pm$ 2.0 & 97.3 $\pm$ 0.8 & 95.8 $\pm$ 0.9 & 98.1 $\pm$ 1.1 & \textit{92.7 $\pm$ 1.7} \\
 &  & MaxLogits & 65.2 $\pm$ 19.4 & 65.2 $\pm$ 19.4 & \textit{82.9 $\pm$ 6.6} & \textit{93.6 $\pm$ 1.8} & \textbf{\textit{97.4 $\pm$ 0.7}} & \textit{95.9 $\pm$ 0.7} & \textit{98.3 $\pm$ 1.0} & \textit{92.7 $\pm$ 1.8} \\
 &  & PostMax & 44.0 $\pm$ 18.2 & 44.0 $\pm$ 18.2 & 72.1 $\pm$ 10.2 & 92.6 $\pm$ 1.8 & \textbf{\textit{97.4 $\pm$ 0.7}} & 95.4 $\pm$ 0.9 & 97.5 $\pm$ 1.0 & 92.5 $\pm$ 1.1 \\
 &  & GHOST & 42.0 $\pm$ 33.9 & 42.0 $\pm$ 33.9 & 78.8 $\pm$ 10.0 & 93.5 $\pm$ 1.8 & \textbf{\textit{97.4 $\pm$ 0.7}} & 95.7 $\pm$ 0.8 & 98.1 $\pm$ 1.2 & 92.6 $\pm$ 2.0 \\
\midrule
CIFAR+50 & CE & MSP & 22.9 $\pm$ 0.0 & 31.9 $\pm$ 3.7 & 54.2 $\pm$ 3.9 & 79.4 $\pm$ 1.6 & \textit{97.1 $\pm$ 0.7} & 90.4 $\pm$ 0.7 & 91.6 $\pm$ 0.6 & 84.4 $\pm$ 0.7 \\
 &  & OpenMax & \textit{24.7 $\pm$ 7.2} & 35.4 $\pm$ 5.3 & 58.4 $\pm$ 3.4 & 81.8 $\pm$ 1.3 & \textit{97.1 $\pm$ 0.7} & 91.2 $\pm$ 0.6 & 92.7 $\pm$ 0.5 & 85.5 $\pm$ 0.8 \\
 &  & MaxLogits & 20.4 $\pm$ 6.3 & \textit{38.0 $\pm$ 6.3} & \textit{61.1 $\pm$ 2.8} & \textit{84.3 $\pm$ 1.0} & \textit{97.1 $\pm$ 0.7} & \textit{92.5 $\pm$ 0.4} & \textit{94.5 $\pm$ 0.4} & \textit{86.0 $\pm$ 0.7} \\
 &  & PostMax & 11.4 $\pm$ 1.4 & 24.2 $\pm$ 6.8 & 51.8 $\pm$ 5.1 & 79.5 $\pm$ 1.2 & \textit{97.1 $\pm$ 0.7} & 90.7 $\pm$ 0.4 & 92.3 $\pm$ 0.4 & 84.0 $\pm$ 0.9 \\
 &  & GHOST & 0.3 $\pm$ 0.2 & 1.4 $\pm$ 0.9 & 11.6 $\pm$ 4.2 & 80.8 $\pm$ 0.7 & \textit{97.1 $\pm$ 0.7} & 90.0 $\pm$ 0.3 & 91.8 $\pm$ 0.4 & 85.5 $\pm$ 0.6 \\
\cmidrule{2-11}
 & ARPL & MSP & \textit{30.0 $\pm$ 0.0} & \textit{38.3 $\pm$ 3.3} & 55.3 $\pm$ 5.3 & 80.3 $\pm$ 1.6 & \textit{97.2 $\pm$ 0.6} & 90.6 $\pm$ 0.9 & 91.8 $\pm$ 0.6 & 85.0 $\pm$ 0.9 \\
 &  & OpenMax & 23.3 $\pm$ 8.3 & 36.1 $\pm$ 8.2 & 58.5 $\pm$ 5.0 & 82.1 $\pm$ 1.3 & 97.1 $\pm$ 0.5 & 91.3 $\pm$ 0.8 & 92.8 $\pm$ 0.5 & 85.2 $\pm$ 1.2 \\
 &  & MaxLogits & 23.5 $\pm$ 12.8 & 36.6 $\pm$ 6.6 & \textit{62.1 $\pm$ 4.7} & \textit{85.0 $\pm$ 1.0} & \textit{97.2 $\pm$ 0.6} & \textit{92.7 $\pm$ 0.6} & \textit{94.7 $\pm$ 0.4} & \textit{86.4 $\pm$ 0.8} \\
 &  & PostMax & 10.3 $\pm$ 5.0 & 23.1 $\pm$ 7.2 & 51.8 $\pm$ 4.0 & 80.1 $\pm$ 1.3 & \textit{97.2 $\pm$ 0.6} & 90.9 $\pm$ 0.6 & 92.5 $\pm$ 0.4 & 84.9 $\pm$ 0.7 \\
 &  & GHOST & 0.1 $\pm$ 0.2 & 1.2 $\pm$ 0.9 & 12.0 $\pm$ 5.6 & 81.3 $\pm$ 1.4 & \textit{97.2 $\pm$ 0.6} & 90.2 $\pm$ 0.6 & 92.0 $\pm$ 0.5 & 86.0 $\pm$ 0.6 \\
\cmidrule{2-11}
 & K+1 & MSP & \textit{48.9 $\pm$ 12.6} & \textit{65.4 $\pm$ 4.4} & \textit{83.8 $\pm$ 2.9} & 93.8 $\pm$ 0.9 & \textbf{\textit{97.7 $\pm$ 0.7}} & \textit{96.3 $\pm$ 0.4} & \textbf{\textit{98.4 $\pm$ 0.5}} & \textit{93.6 $\pm$ 1.0} \\
 &  & OpenMax & 40.4 $\pm$ 6.7 & 60.8 $\pm$ 6.4 & 79.3 $\pm$ 2.9 & 91.6 $\pm$ 1.2 & 97.4 $\pm$ 0.7 & 95.2 $\pm$ 0.7 & 96.9 $\pm$ 0.4 & 91.8 $\pm$ 1.0 \\
 &  & MaxLogits & 38.9 $\pm$ 6.6 & 60.2 $\pm$ 6.9 & 79.8 $\pm$ 2.3 & 93.0 $\pm$ 0.5 & \textbf{\textit{97.7 $\pm$ 0.7}} & 95.9 $\pm$ 0.4 & 97.8 $\pm$ 0.4 & 92.3 $\pm$ 0.6 \\
 &  & PostMax & 41.9 $\pm$ 17.6 & 60.5 $\pm$ 4.2 & 81.8 $\pm$ 3.0 & \textit{93.9 $\pm$ 0.7} & \textbf{\textit{97.7 $\pm$ 0.7}} & 96.2 $\pm$ 0.4 & 98.1 $\pm$ 0.5 & 93.1 $\pm$ 0.8 \\
 &  & GHOST & 5.5 $\pm$ 5.3 & 36.1 $\pm$ 14.8 & 80.8 $\pm$ 2.7 & 93.5 $\pm$ 0.6 & \textbf{\textit{97.7 $\pm$ 0.7}} & 96.0 $\pm$ 0.5 & 97.9 $\pm$ 0.4 & 92.8 $\pm$ 0.5 \\
\cmidrule{2-11}
 & OE & MSP & 50.1 $\pm$ 14.4 & 66.7 $\pm$ 7.6 & 83.5 $\pm$ 3.4 & 94.0 $\pm$ 1.1 & 97.6 $\pm$ 0.4 & 96.2 $\pm$ 0.3 & 98.2 $\pm$ 0.6 & 93.4 $\pm$ 1.0 \\
 &  & OpenMax & 49.6 $\pm$ 13.1 & \textit{67.8 $\pm$ 6.7} & 83.5 $\pm$ 3.5 & 94.1 $\pm$ 1.0 & \textbf{\textit{97.7 $\pm$ 0.5}} & \textbf{\textit{96.4 $\pm$ 0.3}} & 98.2 $\pm$ 0.5 & \textit{93.6 $\pm$ 0.9} \\
 &  & MaxLogits & \textit{50.4 $\pm$ 14.3} & 67.1 $\pm$ 7.7 & \textit{83.7 $\pm$ 3.7} & \textbf{\textit{94.2 $\pm$ 1.0}} & 97.6 $\pm$ 0.4 & 96.2 $\pm$ 0.3 & \textit{98.3 $\pm$ 0.6} & \textit{93.6 $\pm$ 1.0} \\
 &  & PostMax & 46.8 $\pm$ 17.4 & 67.0 $\pm$ 6.8 & 83.2 $\pm$ 3.5 & 94.0 $\pm$ 1.1 & 97.6 $\pm$ 0.4 & 96.2 $\pm$ 0.3 & 98.2 $\pm$ 0.6 & 93.4 $\pm$ 0.9 \\
 &  & GHOST & 7.7 $\pm$ 5.2 & 46.8 $\pm$ 21.1 & 83.4 $\pm$ 3.6 & 94.1 $\pm$ 1.0 & 97.6 $\pm$ 0.4 & 96.2 $\pm$ 0.3 & 98.2 $\pm$ 0.6 & \textit{93.6 $\pm$ 0.9} \\
\cmidrule{2-11}
 & OS & MSP & 51.2 $\pm$ 11.8 & 70.2 $\pm$ 8.2 & 85.1 $\pm$ 2.5 & 94.0 $\pm$ 1.1 & \textit{97.4 $\pm$ 0.5} & \textit{96.2 $\pm$ 0.4} & \textit{98.3 $\pm$ 0.5} & 93.9 $\pm$ 0.8 \\
 &  & OpenMax & 48.0 $\pm$ 11.8 & 70.1 $\pm$ 8.8 & 85.1 $\pm$ 2.6 & 94.0 $\pm$ 1.1 & \textit{97.4 $\pm$ 0.5} & \textit{96.2 $\pm$ 0.4} & \textit{98.3 $\pm$ 0.5} & 93.8 $\pm$ 0.8 \\
 &  & MaxLogits & \textbf{\textit{51.6 $\pm$ 11.9}} & \textbf{\textit{70.4 $\pm$ 8.4}} & \textbf{\textit{85.5 $\pm$ 2.6}} & \textit{94.1 $\pm$ 1.0} & \textit{97.4 $\pm$ 0.5} & \textit{96.2 $\pm$ 0.4} & \textit{98.3 $\pm$ 0.5} & 93.9 $\pm$ 0.8 \\
 &  & PostMax & 35.9 $\pm$ 19.7 & 53.5 $\pm$ 8.1 & 78.2 $\pm$ 4.2 & 92.7 $\pm$ 1.4 & \textit{97.4 $\pm$ 0.5} & 95.6 $\pm$ 0.6 & 97.5 $\pm$ 0.6 & 92.5 $\pm$ 0.7 \\
 &  & GHOST & 8.2 $\pm$ 9.9 & 56.8 $\pm$ 28.2 & \textbf{\textit{85.5 $\pm$ 3.0}} & \textit{94.1 $\pm$ 1.1} & \textit{97.4 $\pm$ 0.5} & \textit{96.2 $\pm$ 0.4} & \textit{98.3 $\pm$ 0.6} & \textbf{\textit{94.0 $\pm$ 0.8}} \\
\bottomrule
\end{tabular}}
  \Caption[tab:summary-table-small]{Small-Scale Evaluation}{
    This table includes a performance overview of all representation learning and postprocessing methods on the small-scale protocols.
    Metrics include CCR@$\fprval$ for $\fprval\in{10^{-2},10^{-1},10^{-0},10^{1},10^{2}}$, AUOSCR, AUROC, and OOSA.
    All scores are reported in percent and CCR@100 is the closed-set accuracy.
    Performance metrics are reported as mean $\pm$ standard deviation over 5 randomized trials.
    The best performing combination for each protocol and metric w.r.t mean score is highlighted in bold.
    The best performing postprocessor for each representation learning method and metric w.r.t mean score is highlighted in italic.
  }
\end{table*}

\begin{table*}[p]
  \centering
  \scalebox{0.54}{\begin{tabular}{@{}l@{}c*{8}{c@\ }c@{}}
\toprule
Dataset & $f$ & $g$ & CCR@0.01 & CCR@0.1 & CCR@1 & CCR@10 & CCR@100 & AUOSCR & AUROC & OOSA \\
\midrule
P1 & CE & MSP & 7.8 & 22.7 & 53.4 & 71.8 & \textit{77.4} & 75.6 & 95.3 & 83.7 \\
 &  & OpenMax & 13.2 & 34.4 & 65.0 & 76.6 & 77.0 & 76.5 & 98.7 & 86.2 \\
 &  & MaxLogits & 6.1 & 31.3 & 64.1 & 76.6 & \textit{77.4} & 76.8 & 98.5 & 86.3 \\
 &  & PostMax & 14.0 & 33.0 & 65.9 & 75.9 & \textit{77.4} & 76.7 & 98.0 & 86.6 \\
 &  & GHOST & \textit{18.3} & \textit{44.9} & \textit{71.4} & \textit{77.0} & \textit{77.4} & \textit{77.1} & \textit{99.1} & \textit{87.5} \\
\cmidrule{2-11}
 & ARPL & MSP & 10.6 & 27.2 & 54.8 & 72.5 & \textit{78.2} & 76.4 & 95.6 & 83.6 \\
 &  & OpenMax & 8.0 & 44.6 & 66.1 & 77.0 & 77.3 & 76.8 & 98.8 & 86.1 \\
 &  & MaxLogits & \textbf{\textit{27.8}} & 35.1 & 66.3 & 77.4 & \textit{78.2} & 77.7 & 98.7 & 86.8 \\
 &  & PostMax & 20.6 & 41.5 & 65.6 & 76.8 & \textit{78.2} & 77.5 & 98.1 & 86.5 \\
 &  & GHOST & 21.1 & \textit{48.9} & \textit{72.3} & \textit{77.9} & \textit{78.2} & \textit{77.9} & \textit{99.2} & \textit{87.7} \\
\cmidrule{2-11}
 & K+1 & MSP & 11.6 & 32.2 & 62.2 & 77.4 & \textbf{\textit{78.4}} & 77.5 & 97.9 & 79.0 \\
 &  & OpenMax & 16.2 & 41.5 & 65.5 & 77.8 & 78.3 & 77.7 & 98.7 & 87.4 \\
 &  & MaxLogits & 17.1 & 39.4 & 72.6 & \textbf{\textit{78.1}} & \textbf{\textit{78.4}} & 78.1 & 99.3 & 88.7 \\
 &  & PostMax & 14.5 & 35.9 & 67.3 & 77.0 & \textbf{\textit{78.4}} & 77.7 & 98.4 & 86.9 \\
 &  & GHOST & \textit{19.7} & \textit{51.3} & \textit{74.8} & \textbf{\textit{78.1}} & \textbf{\textit{78.4}} & \textbf{\textit{78.2}} & \textbf{\textit{99.5}} & \textbf{\textit{89.0}} \\
\cmidrule{2-11}
 & OE & MSP & 5.6 & 27.9 & 62.7 & 77.6 & \textit{78.1} & 77.4 & 98.2 & 84.3 \\
 &  & OpenMax & 5.6 & 28.2 & 62.4 & 77.4 & 78.0 & 77.3 & 98.2 & 83.7 \\
 &  & MaxLogits & \textit{16.1} & 36.7 & 73.1 & 77.7 & \textit{78.1} & \textit{77.8} & 99.0 & 84.3 \\
 &  & PostMax & 8.8 & 27.2 & 63.3 & 77.1 & \textit{78.1} & 77.4 & 98.2 & 84.2 \\
 &  & GHOST & 12.1 & \textit{48.0} & \textit{74.4} & \textit{77.8} & \textit{78.1} & \textit{77.8} & \textit{99.1} & \textit{86.4} \\
\cmidrule{2-11}
 & OS & MSP & 18.0 & 34.2 & 64.9 & 77.8 & \textit{78.3} & 77.7 & 98.6 & 83.4 \\
 &  & OpenMax & \textit{22.5} & 33.4 & 64.6 & 77.6 & 78.0 & 77.4 & 98.6 & 84.3 \\
 &  & MaxLogits & 16.4 & 46.0 & 73.9 & \textit{78.0} & \textit{78.3} & \textit{78.0} & 99.3 & 86.4 \\
 &  & PostMax & 21.4 & 41.4 & 70.7 & \textit{78.0} & \textit{78.3} & 77.9 & 99.1 & \textit{87.3} \\
 &  & GHOST & 19.5 & \textbf{\textit{54.5}} & \textbf{\textit{75.9}} & \textit{78.0} & \textit{78.3} & \textit{78.0} & \textit{99.4} & 83.9 \\
\midrule
P2 & CE & MSP & --- & --- & 21.8 & 50.5 & \textit{74.0} & 64.9 & 80.1 & 76.4 \\
 &  & OpenMax & 2.5 & 3.6 & 24.5 & 51.5 & 73.7 & 66.3 & 83.4 & 77.1 \\
 &  & MaxLogits & 7.6 & 8.4 & 25.0 & 58.9 & \textit{74.0} & 68.5 & 87.8 & 78.8 \\
 &  & PostMax & 6.0 & 7.4 & \textit{30.5} & 59.9 & \textit{74.0} & 68.8 & 88.2 & 79.4 \\
 &  & GHOST & \textit{8.1} & \textit{10.3} & 30.3 & \textit{61.3} & \textit{74.0} & \textit{69.1} & \textit{88.9} & \textit{80.0} \\
\cmidrule{2-11}
 & ARPL & MSP & --- & --- & 24.4 & 49.9 & \textit{75.4} & 66.1 & 80.7 & 76.3 \\
 &  & OpenMax & 2.6 & 4.1 & 29.5 & 52.2 & 73.8 & 66.7 & 83.6 & 77.2 \\
 &  & MaxLogits & 3.1 & 3.7 & 26.9 & 58.5 & \textit{75.4} & 69.2 & 87.5 & 78.8 \\
 &  & PostMax & \textit{5.7} & \textbf{\textit{13.6}} & \textit{34.5} & 59.8 & \textit{75.4} & 69.8 & 88.3 & 80.2 \\
 &  & GHOST & 4.5 & 13.3 & \textit{34.5} & \textit{60.1} & \textit{75.4} & \textit{70.1} & \textit{88.9} & \textbf{\textit{80.5}} \\
\cmidrule{2-11}
 & K+1 & MSP & --- & --- & 25.8 & 55.8 & \textbf{\textit{76.9}} & 68.5 & 84.5 & 65.2 \\
 &  & OpenMax & 3.5 & 5.9 & 33.6 & 59.1 & 76.4 & 70.5 & 87.2 & 78.8 \\
 &  & MaxLogits & 2.3 & 4.3 & 33.0 & 62.5 & \textbf{\textit{76.9}} & 71.5 & 89.2 & \textit{79.7} \\
 &  & PostMax & 8.3 & \textit{11.0} & 34.5 & 63.4 & \textbf{\textit{76.9}} & 71.8 & 89.4 & 79.2 \\
 &  & GHOST & \textbf{\textit{10.1}} & 10.6 & \textbf{\textit{40.1}} & \textbf{\textit{64.4}} & \textbf{\textit{76.9}} & \textbf{\textit{72.2}} & \textbf{\textit{90.3}} & 79.0 \\
\cmidrule{2-11}
 & OE & MSP & --- & 5.0 & 20.9 & 54.8 & 74.3 & 66.9 & 84.5 & 71.6 \\
 &  & OpenMax & --- & 3.3 & 21.5 & 55.1 & \textit{74.5} & 67.1 & 84.6 & 72.2 \\
 &  & MaxLogits & 2.5 & 2.7 & 23.1 & 55.7 & 74.3 & 66.9 & 85.1 & 71.0 \\
 &  & PostMax & 2.3 & 8.3 & 24.5 & 57.8 & 74.3 & \textit{67.9} & \textit{85.9} & \textit{73.3} \\
 &  & GHOST & \textit{4.7} & \textit{10.3} & \textit{27.3} & \textit{57.9} & 74.3 & 67.6 & \textit{85.9} & 71.5 \\
\cmidrule{2-11}
 & OS & MSP & 3.1 & 4.4 & 25.1 & 56.5 & \textit{74.9} & 67.8 & 85.1 & 71.2 \\
 &  & OpenMax & 4.2 & 4.7 & 25.7 & 57.1 & \textit{74.9} & 67.9 & 85.3 & 71.1 \\
 &  & MaxLogits & 6.3 & 6.5 & 26.1 & 58.1 & \textit{74.9} & 68.0 & 85.9 & 71.2 \\
 &  & PostMax & 5.8 & 9.3 & 27.9 & \textit{58.6} & \textit{74.9} & \textit{68.4} & 85.8 & \textit{73.9} \\
 &  & GHOST & \textit{7.5} & \textit{10.0} & \textit{32.3} & 58.5 & \textit{74.9} & 68.1 & \textit{86.0} & 70.2 \\
\midrule
P3 & CE & MSP & --- & --- & \textit{21.4} & 64.9 & \textit{85.7} & 78.1 & 86.4 & 78.2 \\
 &  & OpenMax & --- & --- & 20.9 & 64.6 & 85.3 & 77.9 & 86.4 & 78.0 \\
 &  & MaxLogits & \textit{1.2} & \textit{3.2} & 14.6 & 62.6 & \textit{85.7} & 77.9 & 87.7 & 78.1 \\
 &  & PostMax & 0.1 & 1.9 & 17.7 & \textit{66.3} & \textit{85.7} & \textit{78.9} & \textit{88.7} & \textit{79.1} \\
 &  & GHOST & 0.4 & 2.3 & 20.1 & 65.9 & \textit{85.7} & 78.8 & 88.6 & 78.7 \\
\cmidrule{2-11}
 & ARPL & MSP & --- & \textbf{\textit{6.8}} & 21.5 & 64.5 & \textbf{\textit{86.5}} & 78.4 & 86.0 & 78.0 \\
 &  & OpenMax & --- & \textbf{\textit{6.8}} & \textit{21.7} & 64.4 & 86.3 & 78.3 & 86.0 & 78.0 \\
 &  & MaxLogits & 0.5 & 2.5 & 15.8 & 62.0 & \textbf{\textit{86.5}} & 78.4 & 87.6 & 77.9 \\
 &  & PostMax & 1.4 & 4.8 & 17.4 & \textit{65.4} & \textbf{\textit{86.5}} & \textit{79.3} & \textit{88.6} & \textit{79.0} \\
 &  & GHOST & \textbf{\textit{2.3}} & 3.4 & 19.5 & 65.0 & \textbf{\textit{86.5}} & 79.2 & 88.5 & 78.6 \\
\cmidrule{2-11}
 & K+1 & MSP & --- & 3.7 & 24.5 & 65.6 & \textit{85.6} & 77.8 & 86.5 & 70.7 \\
 &  & OpenMax & --- & --- & 22.5 & 65.6 & \textit{85.6} & 78.5 & 87.0 & 77.1 \\
 &  & MaxLogits & 1.0 & 3.2 & 19.1 & 66.0 & \textit{85.6} & 78.7 & 88.5 & 77.5 \\
 &  & PostMax & \textit{2.1} & 4.2 & 23.5 & \textbf{\textit{69.1}} & \textit{85.6} & \textbf{\textit{79.7}} & \textbf{\textit{89.5}} & \textbf{\textit{79.4}} \\
 &  & GHOST & 0.5 & \textit{4.7} & \textit{25.4} & 68.8 & \textit{85.6} & 79.6 & 89.4 & 78.5 \\
\cmidrule{2-11}
 & OE & MSP & --- & --- & 21.8 & 66.9 & 84.6 & 77.7 & 87.5 & 72.6 \\
 &  & OpenMax & --- & --- & 21.7 & 66.7 & \textit{84.7} & \textit{77.8} & 87.5 & 72.8 \\
 &  & MaxLogits & \textit{1.1} & \textit{2.7} & 17.7 & 65.0 & 84.6 & 77.1 & 87.2 & 72.7 \\
 &  & PostMax & 0.3 & 1.5 & 12.7 & 61.5 & 84.6 & 77.0 & 86.8 & \textit{75.5} \\
 &  & GHOST & 0.1 & 2.6 & \textit{22.6} & \textit{67.3} & 84.6 & \textit{77.8} & \textit{87.8} & 73.8 \\
\cmidrule{2-11}
 & OS & MSP & --- & --- & \textbf{\textit{26.6}} & 67.4 & \textit{84.9} & 78.1 & 87.8 & 75.2 \\
 &  & OpenMax & --- & --- & 26.1 & 67.2 & 84.8 & 78.0 & 87.8 & \textit{75.4} \\
 &  & MaxLogits & 0.8 & 2.7 & 20.1 & 66.9 & \textit{84.9} & 77.6 & 87.6 & 74.7 \\
 &  & PostMax & --- & 1.4 & 11.4 & 60.6 & \textit{84.9} & 76.9 & 86.3 & 71.4 \\
 &  & GHOST & \textit{1.1} & \textit{4.8} & 25.2 & \textit{68.5} & \textit{84.9} & \textit{78.3} & \textit{88.4} & 74.9 \\
\bottomrule
\end{tabular}}
  \Caption[tab:summary-table-large]{Large-Scale Evaluation}{
    This table repeats the performance overview of \tab{summary-table-small} on large-scale protocols.
    --- indicates that at no threshold $\theta_\fprval$ for corresponding FPR $\fprval$ could be achieved.
  }
\end{table*}

\begin{table}
  \centering
  \scalebox{0.93}{\begin{tabular}{l|rr|rr|rr}
\toprule
Model & \multicolumn{2}{c|}{Class 1} & \multicolumn{2}{c|}{Class 9} & \multicolumn{2}{c}{Knowns Except 1 and 9} \\ 
 & min & max & min & max & min & max \\
\midrule
CE & 94.1 & 97.6 & 89.5 & 91.9 & 86.7 & 97.9 \\
ARPL & 86.7 & 96.9 & 89.5 & 92.2 & 83.8 & 98.0 \\
K+1 & 67.0 & 94.8 & 92.4 & 95.6 & 95.2 & 99.5 \\
OE & 65.7 & 86.6 & 94.2 & 97.2 & 96.7 & 99.8 \\
OS & 58.1 & 88.7 & 93.6 & 95.3 & 97.7 & 99.6 \\
\bottomrule
\end{tabular}
}
  \Caption[tab:emnist-summary-table]{Class-Wise CCR}{EMNIST ranges of class-wise CCR are given at the operational threshold, $\mathrm{CCR}_c(\threshold^\star)$, for classes 1, 9, and rest (all other known classes) per representation learning method.}
\end{table}

\begin{figure}[p!]
  \centering
  \includegraphics[width=.87\textheight, angle=90]{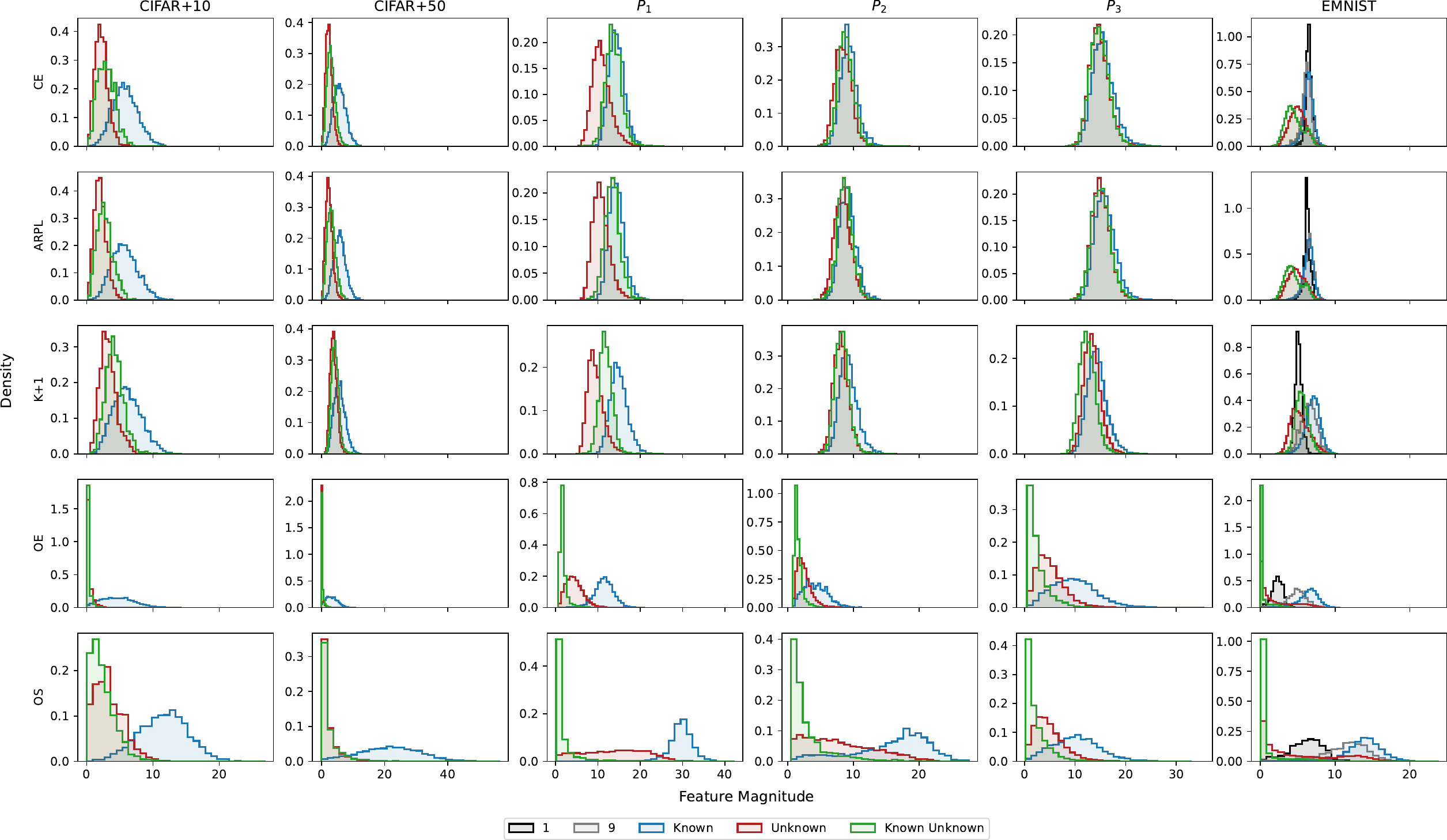}
  \Caption[fig:magnitudes-complete]{Magnitudes}{
    Feature magnitude distributions of known, known unknown, and unknown classes on all protocols and for all representation learning methods, complete version of \fig{magnitudes}.
    OE and OS experience feature magnitude collapse on \p3 and EMNIST, pulling their feature magnitudes towards zero.
    For EMNIST, we show distributions for known classes 1 and 9 (black and grey) that are highly similar to known unknown classes, and other known classes (blue) separately.
  }
\end{figure}

\begin{figure*}[t]
  \centering
  \begin{subfigure}{.99\textwidth}
    \centering
    \includegraphics[width=\textwidth]{AUOSCR_heatmaps.pdf}
    \caption{AUOSCR}
    \label{fig:supplemental-auoscr-heatmaps}
  \end{subfigure}
  \begin{subfigure}{.99\textwidth}
    \centering
    \includegraphics[width=\textwidth]{OOSA_heatmaps.pdf}
    \caption{OOSA}
    \label{fig:supplemental-oosa-heatmaps}
  \end{subfigure}
  \begin{subfigure}{.99\textwidth}
    \centering
    \includegraphics[width=\textwidth]{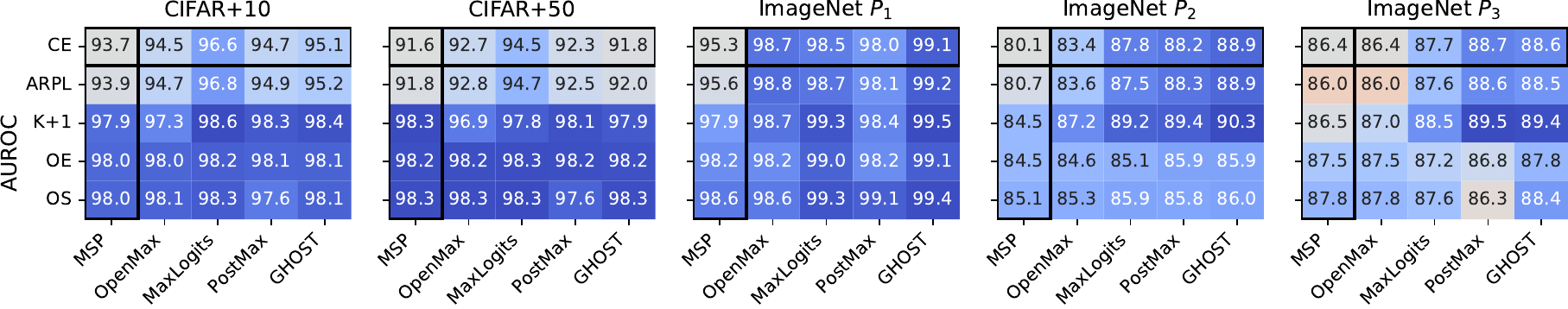}
    \caption{AUROC}
    \label{fig:supplemental-auroc-heatmaps}
  \end{subfigure}
  \begin{subfigure}{.99\textwidth}
    \centering
    \includegraphics[width=\textwidth]{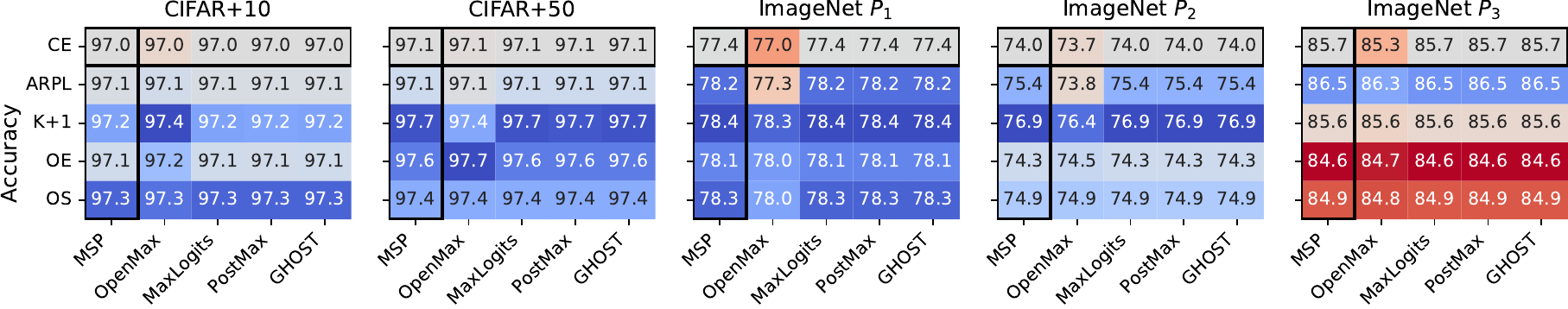}
    \caption{Accuracy}
    \label{fig:supplemental-accuracy-heatmaps}
  \end{subfigure}
  \Caption[fig:supplemental-score-heatmaps]{Absolute Performance}{These heatmaps show the absolute values for evaluations of \subref{fig:supplemental-auoscr-heatmaps} AUOSCR, \subref{fig:supplemental-oosa-heatmaps} OOSA, \subref{fig:supplemental-auroc-heatmaps} AUROC, and \subref{fig:supplemental-accuracy-heatmaps} Accuracy. Each heatmap is normalized independently and centered around CE+MSP, where blue shows an increase and red a decrease. Baseline representation learning and postprocessing methods are surrounded by a black border. Results for CIFAR+N are averaged over 5 trials.}
\end{figure*}

\begin{figure*}[t]
  \centering
  \begin{subfigure}{.99\textwidth}
    \centering
    \includegraphics[width=\textwidth, clip, trim=0 25 0 0]{AUOSCR_delta_scatter.pdf}
    \caption{AUOSCR}
    \label{fig:supplemental-auoscr-delta-scatter}
  \end{subfigure}
  \begin{subfigure}{.99\textwidth}
    \centering
    \includegraphics[width=\textwidth, clip, trim=0 25 0 0]{OOSA_delta_scatter.pdf}
    \caption{OOSA}
    \label{fig:supplemental-oosa-delta-scatter}
  \end{subfigure}
  \begin{subfigure}{.99\textwidth}
    \centering
    \includegraphics[width=\textwidth, clip, trim=0 25 0 0]{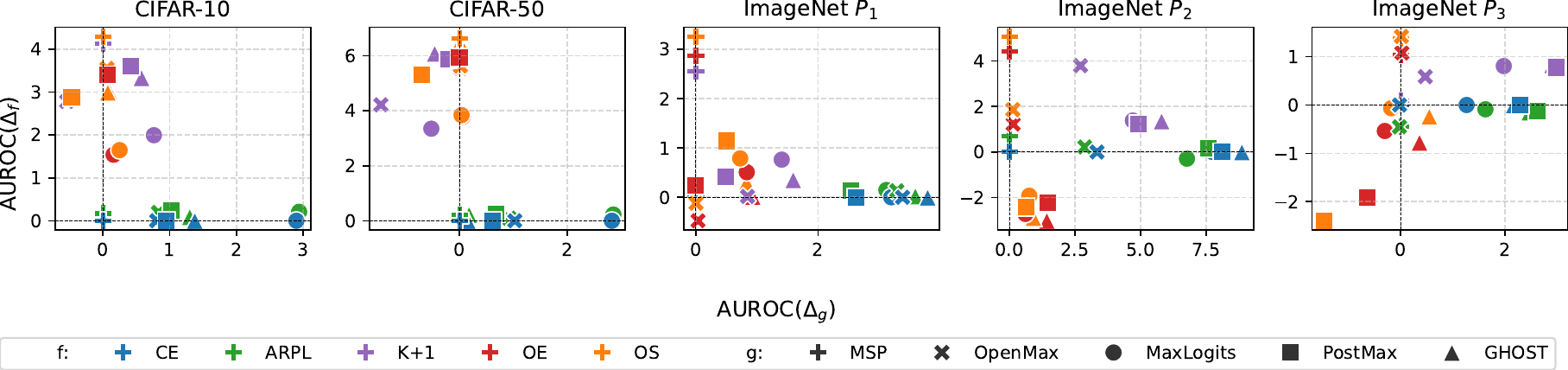}
    \caption{AUROC}
    \label{fig:supplemental-auroc-delta-scatter}
  \end{subfigure}
  \begin{subfigure}{.99\textwidth}
    \centering
    \includegraphics[width=\textwidth]{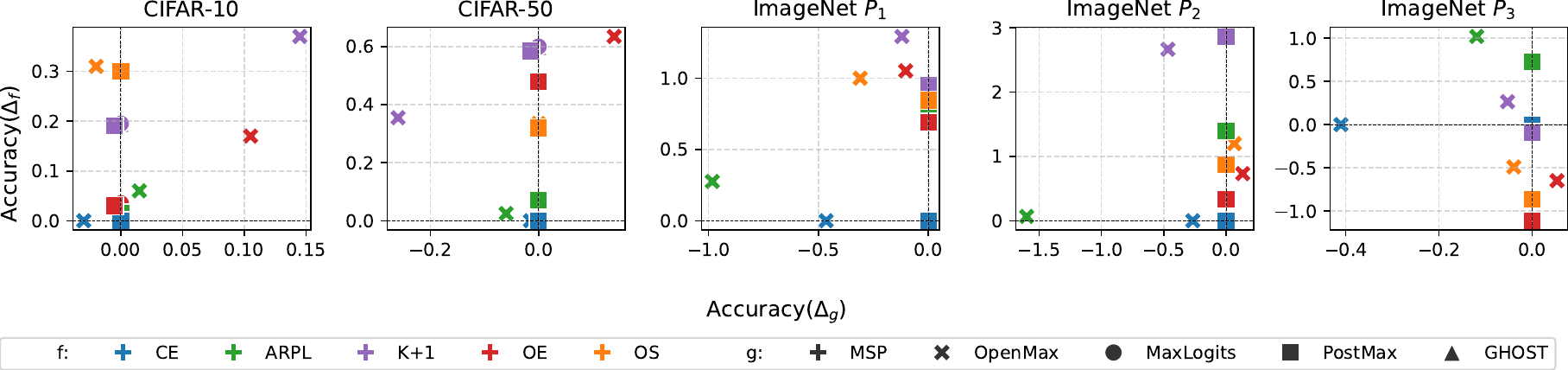}
    \caption{Accuracy}
    \label{fig:supplemental-accuracy-delta-scatter}
  \end{subfigure}
  \Caption[fig:supplemental-delta-scatter]{Interaction Effects}{
    This figure shows the interaction effects of representation learning and postprocessing components as correlation between representation learning performance contribution \deltarl{} and postprocessor contribution \deltapp{} in terms of \subref{fig:supplemental-auoscr-delta-scatter} AUOSCR, \subref{fig:supplemental-oosa-delta-scatter} OOSA, \subref{fig:supplemental-auroc-delta-scatter} AUROC, and \subref{fig:supplemental-accuracy-delta-scatter} Accuracy.
    Results for CIFAR+N are averaged over 5 trials.
  }
\end{figure*}

\begin{figure*}
  \centering
  \begin{subfigure}{.58\textwidth}
    \includegraphics[width=.99\textwidth, clip, trim=0 45 0 0]{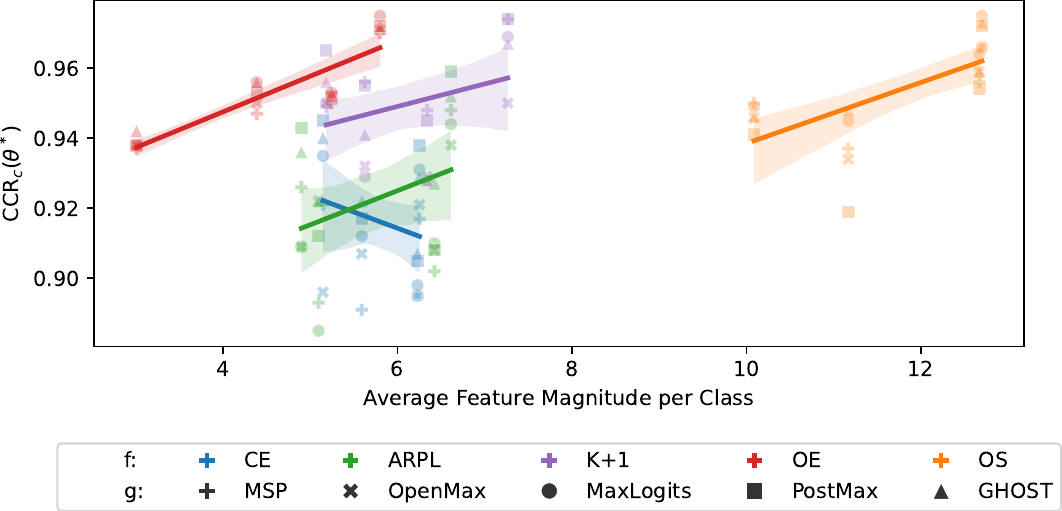}
    \caption{CIFAR+10}
    \label{fig:supplemental-regression-cifar10}
  \end{subfigure}
  \begin{subfigure}{.58\textwidth}
    \includegraphics[width=.99\textwidth, clip, trim=0 45 0 0]{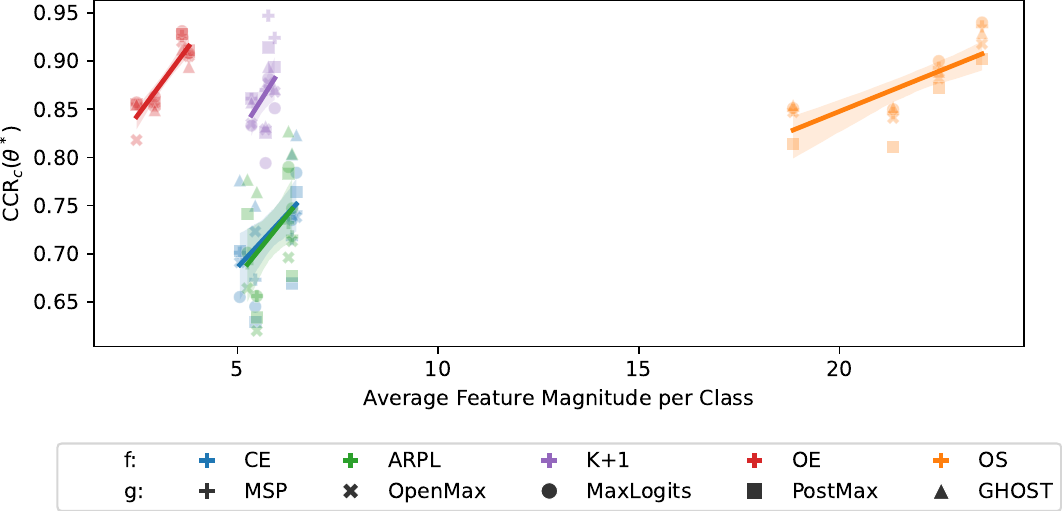}
    \caption{CIFAR+50}
    \label{fig:supplemental-regression-cifar50}
  \end{subfigure}
  \begin{subfigure}{.58\textwidth}
    \includegraphics[width=.99\textwidth, clip, trim=0 45 0 0]{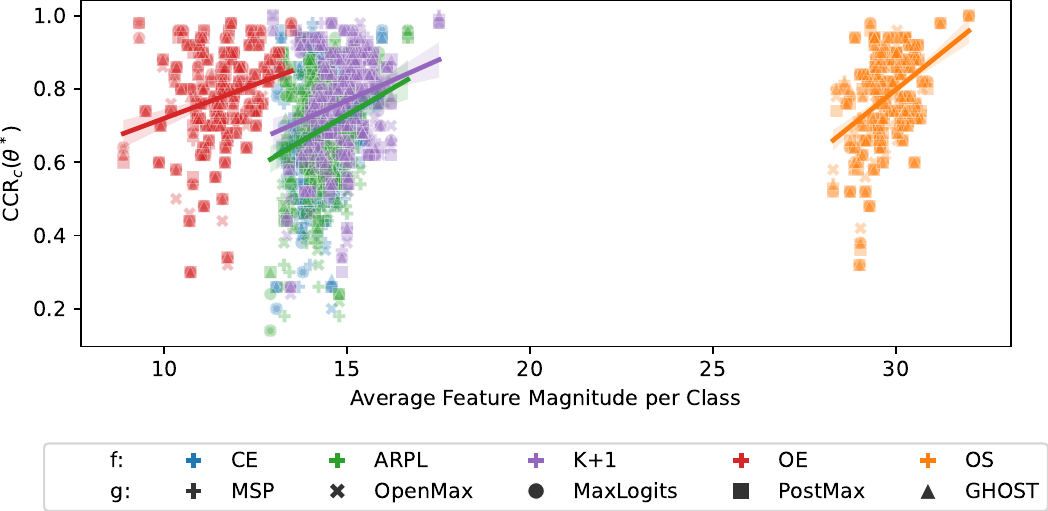}
    \caption{\p1}
    \label{fig:supplemental-regression-p1}
  \end{subfigure}
  \begin{subfigure}{.58\textwidth}
    \includegraphics[width=.99\textwidth, clip, trim=0 45 0 0]{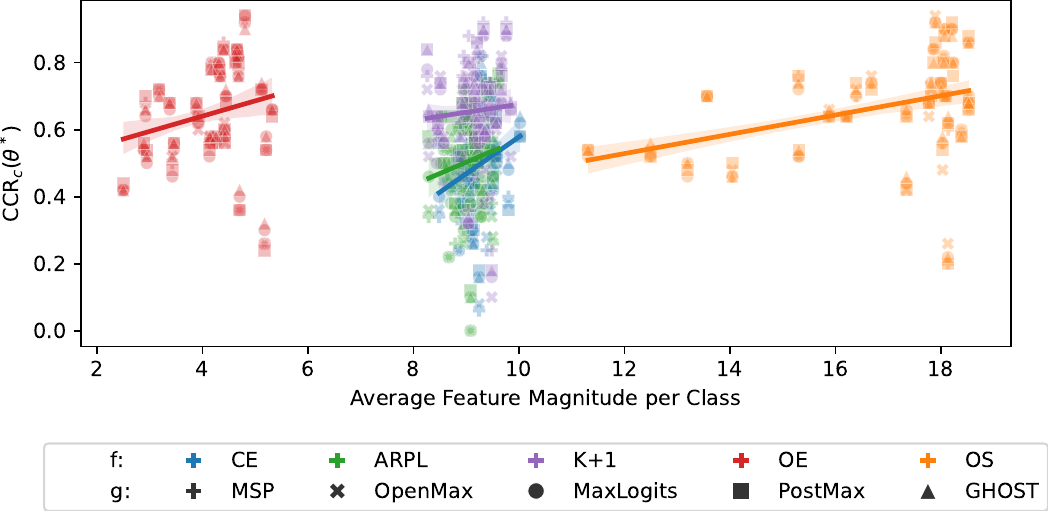}
    \caption{\p2}
    \label{fig:supplemental-regression-p2}
  \end{subfigure}
  \begin{subfigure}{.58\textwidth}
    \includegraphics[width=.99\textwidth]{classwise_CCR_vs_feature_magnitude_P3_regression.pdf}
    \caption{\p3}
    \label{fig:supplemental-regression-p3}
  \end{subfigure}
  \Caption[fig:supplemental-regressions]{Regression Analysis}{
    Linear regressions of class-wise CCR at the operational threshold, $\mathrm{CCR}_c(\threshold^*)$, against class-wise average feature magnitude for known classes.
  Regression is performed for each representation learning method independently and over all postprocessors. For CIFAR+N results are reported for the first trial only.}
\end{figure*}

\end{document}